\documentclass[10pt,journal,compsoc]{IEEEtran}

\ifCLASSOPTIONcompsoc
  \usepackage[nocompress]{cite}
\else
  \usepackage{cite}
\fi

\ifCLASSINFOpdf
  \usepackage[pdftex]{graphicx}
  \graphicspath{{../figures/}}
  \DeclareGraphicsExtensions{.pdf}
\else
  \usepackage[dvips]{graphicx}
  \graphicspath{{../figures/}}
  \DeclareGraphicsExtensions{.pdf}
\fi

\usepackage{amsmath}
\usepackage{amssymb}
\usepackage{array}
\usepackage{subfig}
\usepackage{booktabs}
\usepackage{algorithm}  
\usepackage{algorithmicx}
\usepackage{algpseudocode}  
\usepackage{multirow}
\usepackage{enumerate}
\usepackage{xcolor}
\usepackage{colortbl}
\usepackage{comment}
\usepackage{graphicx}
\usepackage{amssymb}
\usepackage[pagebackref,breaklinks,colorlinks]{hyperref}

\usepackage[capitalize]{cleveref}
\crefname{section}{Sec.}{Secs.}
\Crefname{section}{Section}{Sections}
\Crefname{table}{Table}{Tables}
\crefname{table}{Tab.}{Tabs.}

\newcommand{\revise}[1]{\textcolor{black}{#1}}

\begin{document}
\title{MOODv2: Masked Image Modeling for Out-of-Distribution Detection}

% Jingyao Li, Pengguang Chen, Shaozuo Yu, Shu Liu, Jiaya Jia
\author{Jingyao~Li,
        Pengguang~Chen,
        Shaozuo~Yu,
        Shu~Liu,~\IEEEmembership{Member,~IEEE}
        and~Jiaya~Jia,~\IEEEmembership{Fellow,~IEEE}% <-this % stops a space
\IEEEcompsocitemizethanks{\IEEEcompsocthanksitem Jingyao Li and Shaozuo Yu are with the Department of Computer Science and Engineering of the Chinese University of Hong Kong (CUHK) \\
% note need leading \protect in front of \\ to get a newline within \thanks as
% \\ is fragile and will error, could use \hfil\break instead.
Jiaya Jia's E-mail: leojia9@gmail.com
\IEEEcompsocthanksitem Pengguang Chen, Shu Liu and Jiaya Jia are with SmartMore.}% <-this % stops an unwanted space
}

\IEEEtitleabstractindextext{%
\begin{abstract}
\revise{The crux of effective out-of-distribution (OOD) detection lies in acquiring a robust in-distribution (ID) representation, distinct from OOD samples. While previous methods predominantly leaned on recognition-based techniques for this purpose, they often resulted in shortcut learning, lacking comprehensive representations. In our study, we conducted a comprehensive analysis, exploring distinct pretraining tasks and employing various OOD score functions. The results highlight that the feature representations pre-trained through reconstruction yield a notable enhancement and narrow the performance gap among various score functions. This suggests that even simple score functions can rival complex ones when leveraging reconstruction-based pretext tasks. Reconstruction-based pretext tasks adapt well to various score functions. As such, it holds promising potential for further expansion. Our OOD detection framework, MOODv2, employs the masked image modeling pretext task. Without bells and whistles, MOODv2 impressively enhances 14.30\% AUROC to 95.68\% on ImageNet and achieves 99.98\% on CIFAR-10.}
\end{abstract}

% Note that keywords are not normally used for peerreview papers.

% Codes are available at \href{}{https://github.com/JulietLJY/MOOD}
}

\maketitle
\IEEEdisplaynontitleabstractindextext
\IEEEpeerreviewmaketitle

\IEEEraisesectionheading{\section{Introduction}\label{sec:intro}}
\IEEEPARstart{A} reliable visual recognition system not only provides correct predictions on known context (also known as in-distribution data) but also detects unknown out-of-distribution (OOD) samples and rejects (or transfers) them to human intervention for safe handling. This motivates the applications of outlier detectors before feeding input to the downstream networks, which is the main task of OOD detection, also referred to as novelty or anomaly detection. OOD detection is the task of identifying whether a test sample is drawn far from the in-distribution (ID) data or not. It is at the cornerstone of various safety-critical applications, including medical diagnosis \cite{caruana2015intelligible}, fraud detection \cite{phua2010comprehensive}, autonomous driving \cite{eykholt2018robust}, etc. \revise{A representative in-distribution feature space representation is crucial for out-of-distribution detection. A well-crafted feature representation significantly enhances the performance via most mainstream OOD detection score functions. Our research is dedicated to refining feature representations tailored for OOD detection, with the aim of advancing the entire field.}

\begin{figure*}[t]
  \centering
    \includegraphics[width=0.9\linewidth, trim=75 225 380 120, clip]{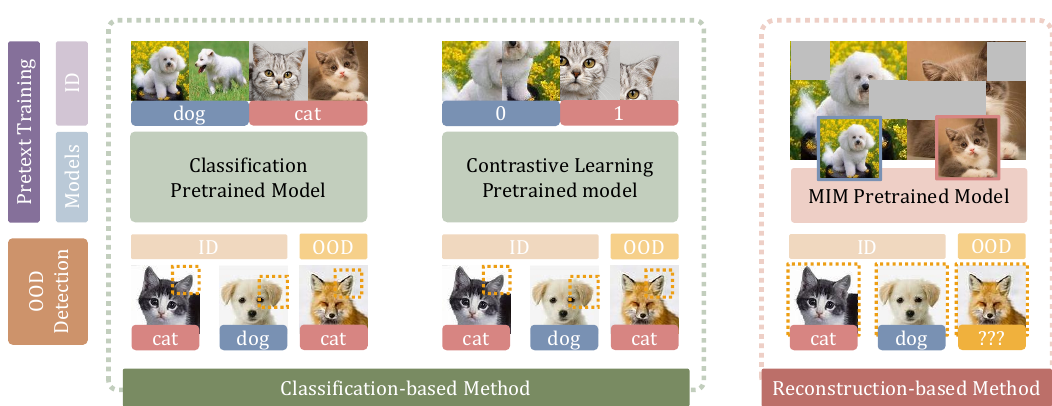}
    \caption{\revise{Comparison of reconstruction-based and classification-based methods. In the context of image classification, networks often take a shortcut when categorizing images \cite{backdoor_attack, frog_attack}. For example, ears are a distinctive feature for distinguishing between cats and dogs, and a classification model typically assumes that animals with pointed ears are cats, while those without are dogs. Consequently, when the network encounters an out-of-distribution animal, such as a fox with pointed ears, it readily misclassifies it as a cat. In contrast, reconstruction-based tasks effectively mitigate this issue. By randomly masking portions of images, the model avoids learning localized, stereotypical features (e.g., masked ears), thus preventing shortcuts and instead acquiring effective pixel-level representations for ID data. This significantly improves the model's ability to detect OOD instances.}}
  \label{fig:intro}
\end{figure*}

Existing methods perform contrastive learning \cite{csi, ssd} or pretrain classification on a large dataset \cite{oodlimits, vim, yang2021generalized, sariyildiz2023fake} to detect OOD samples. The former methods classify images according to the pseudo labels while the latter classifies images based on ground truth, whose core tasks are both to fulfill the classification target. However, research on backdoor attack \cite{backdoor_attack, frog_attack} shows that when learning is represented by classifying data, networks tend to take a shortcut to classify images. In a typical backdoor attack scene \cite{frog_attack}, the attacker adds secret triggers on original training images with the visibly correct label. During the course of testing, the victim model classifies images with secret triggers into the wrong category. Research in this area demonstrates that networks only learn specific distinguishable patterns of different categories because it is a shortcut to fulfill the classification requirement. Nonetheless, learning these patterns is ineffective for OOD detection. Thus, learning representations by classifying ID data for OOD detection may not be satisfying. For example, when patterns similar to some ID categories appear in OOD samples, the network could easily interpret these OOD samples as the ID data and classify them into the wrong ID categories\revise{, as shown in \cref{fig:intro}}. 

\begin{figure}[t]
  \centering
    \includegraphics[width=0.8\linewidth]{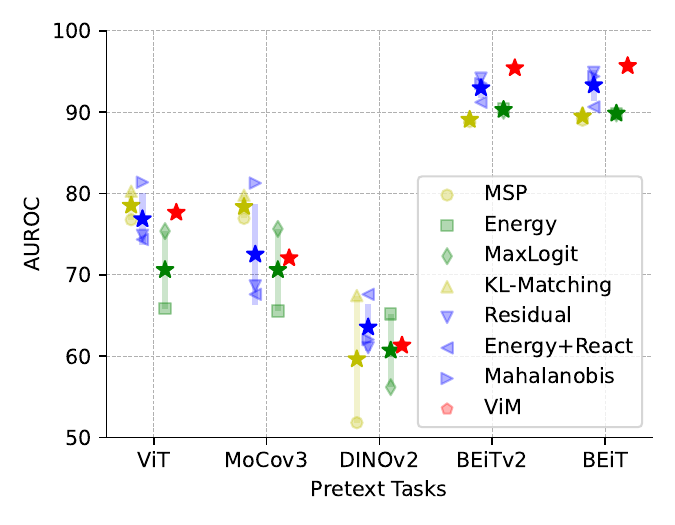}
    \caption{\revise{The average AUROC (\%) tested on four OOD datasets applied to a ViT model with different pre-text tasks. 
    Methods in blue use the feature space;
    methods in green use logits;
    methods in yellow use the softmax probability;
    and methods in red use both features and logits. The stars show the average performance of a category of methods.}}
  \label{fig:ablation}
\end{figure}

\iffalse
\begin{figure}[t]
  \centering
    \includegraphics[width=0.99\linewidth]{figures/impression.pdf}
    \caption{\revise{The AUROC (in percentage) of eight OOD detection algorithms applied to a ViT model with five pre-text tasks.The OOD datasets are ImageNet-O ($x$-axis) and OpenImage-O ($y$-axis).}}
  \label{fig:impression}
\end{figure}
\fi

\revise{To remedy this issue, we introduce the reconstruction-based pretext task. Different from contrastive learning in existing OOD detection approaches \cite{csi, ssd}, our method forces the network to achieve the training purpose of reconstructing the image and thus makes it learn pixel-level feature representation. Specifically, we adopt the masked image modeling (MIM) \cite{beit} as our self-supervised pretext task, which has been demonstrated to have great potential in both natural language processing \cite{bert} and computer vision \cite{beit, mae}. In the MIM task, a proportion of image patches are randomly masked. The network learns information from the remaining patches to speculate the masked patches and restore tokens of the original image. The reconstruction process enables the model to learn from the prior effective ID feature representation rather than just learning different patterns among categories in the classification process. In our work, we observed that the pre-trained models effectively reconstruct ID images, whereas they exhibit distinct domain differences when it comes to the OOD domain (\cref{fig:recover}). This visual discrepancy clearly underscores the existing domain gap in model features between ID and OOD data, offering valuable insights for OOD detection.}

\revise{To validate the effectiveness of our ID feature representation, we conduct experiments to test its performance with various mainstream OOD detection score functions. We employed OOD score functions encompassing probability-based \cite{baseline_ood, maxlogit}, logits-based \cite{energy, maxlogit}, features-based \cite{vim, react, mahalanobis}, and hybrid methods utilizing both logits and features \cite{vim}. In the context of a comparative analysis spanning classic classification \cite{vit}, contrastive learning \cite{mocov3, dinov2}, and masked image modeling pretext tasks \cite{beit, beitv2}, our findings underscore the dominant role of reconstruction-based strategies in the field of OOD detection, as illustrated in \cref{fig:ablation}.}

\revise{Furthermore, we conduct a comprehensive analysis of the experimental results and observe that our approach not only significantly improves the overall results but also substantially reduces the disparities among score functions. This observation underscores that even simple score functions can perform on par with more complex ones when a representative ID feature representation is utilized. These findings further emphasize the critical importance of effective feature representation in OOD detection. More details are in \cref{sec:pretask}.} \revise{Ultimately, MOODv2 demonstrates remarkable enhancements, achieving a substantial 14.30\% increase, reaching 95.68\% AUROC on ImageNet. On CIFAR-10, our results significantly improved to an impressive 99.98\%, marking a notable 0.35\% enhancement compared to the previous state-of-the-art.}

\section{Related Works}\label{sec:related}

\subsection{Out-of-distribution Detection}
\revise{Many scoring functions have been developed by researchers to distinguish between in-distribution and out-of-distribution examples. These functions are designed to exploit properties that are typically exhibited by ID examples but violated by OOD examples, and vice versa. These scores are primarily derived from three sources:}
\begin{enumerate}

    \item \revise{\textbf{Probability-based}: This category includes measures like the maximum softmax probabilities~\cite{baseline_ood} and the minimum KL-divergence between the softmax and the mean class-conditional distributions~\cite{maxlogit}, etc.}

    \item \revise{\textbf{Logit-based}: These functions rely on maximum logits~\cite{maxlogit} and the \(\mathrm{logsumexp}\) function computed over logits~\cite{energy}, etc.}

    \item \revise{\textbf{Feature-based}: These functions involve the norm of the residual between a feature and the pre-image of its low-dimensional embedding~\cite{ndiour2020out} and the minimum Mahalanobis distance between a feature and the class centroids~\cite{mahalanobis}, among others.}

\end{enumerate}
\revise{After a thorough analysis of the performance and their correlations with various score functions and pretext tasks, our work follows the hybrid methods combining logit and feature \cite{vim}} and includes the reconstruction-based methods as a pretext task. We will explain the implementation details later in this paper.

\subsection{Self-Supervised Pretext Task}
\revise{In the ever-evolving landscape of computer vision and deep learning, a multitude of strategies and techniques have been devised to enhance the capacity of models to understand and process visual data:.}

\begin{enumerate}

    \item \revise{\textbf{Classification task:} Vision models are pre-trained via classical classification task \cite{vit}.}

    \item \revise{\textbf{Contrastive Learning tasks: }MOCOv3 \cite{mocov3} and DINOv2 \cite{dinov2} are advanced contrastive learning methods used for self-supervised representation learning. These methods focus on learning representations by contrasting positive pairs (e.g., different augmentations of the same image) with negative pairs (e.g., augmentations from different images). MOCOv3 extends the MOCO framework \cite{moco} with a momentum encoder and dynamic queues for improved performance. DINOv2 introduces a clustered teacher network and an asymmetric loss to learn efficient representations.}

    \item \revise{\textbf{Masked Image Modeling Tasks: } Data-Efficient Image Transformer (BEiT series \cite{beit, beitv2}) are self-supervised learning tasks that involve masked image modeling. In these tasks, a portion of an image is randomly masked, and the model's objective is to predict the masked pixels, effectively filling in the blanks.}
    
\end{enumerate}

\revise{These methods and tasks represent cutting-edge approaches in the field of computer vision and deep learning. They have led to substantial improvements in the ability of models to learn useful visual representations from unlabeled data, enabling better performance on various downstream vision tasks.}

Multiple existing methods take advantage of self-supervised tasks to guide the learning of representation for OOD detection. Previous work \cite{csi, ssd} presents contrastive learning models as feature extractors. However, existing approaches of classifying transformed images according to contrastive learning possess similar limitations -- that is, the model tends to learn the specific patterns of categories \cite{backdoor_attack, backdoor_survey}, which are beneficial for classification but do not help understand the intrinsic ID representation. In our work, we address this issue by performing the masked image modeling task for OOD detection.

\subsection{Training Strategy}
\revise{Numerous approaches have been developed to address OOD-awareness in training loss~\cite{confbranch2018arxiv}. These methods often involve the introduction of regularization terms aimed at encouraging a clearer separation between ID and OOD features~\cite{onedim21cvpr,huang2021mos}. In some cases, networks are augmented with confidence estimation branches, utilizing misclassified in-distribution examples as proxies for out-of-distribution ones~\cite{confbranch2018arxiv}. MOS~\cite{huang2021mos} adapts the loss function by incorporating a predefined group structure, enabling the minimum group-wise ``else" class probability to serve as an indicator of OOD classification. An alternative approach~\cite{onedim21cvpr} focuses on compelling ID samples to embed into a union of 1-dimensional subspaces during training, and it evaluates the minimum angular distance between the feature and class-wise subspaces. }

\revise{In contrast to these approaches, our method belongs to the lightweight training-free methods \cite{vim, MOOD}, which doesn't necessitate retraining the model. Therefore, it not only offers a more straightforward application but also preserves the accuracy of in-distribution classification.}

\begin{figure}[t]
  \centering
    \subfloat[ImageNet]{\includegraphics[width=0.99\linewidth]{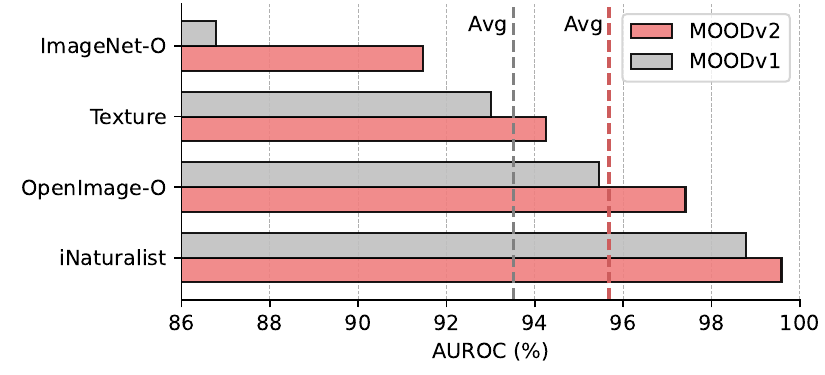}}
    
    \subfloat[CIFAR-10]{\includegraphics[width=0.99\linewidth]{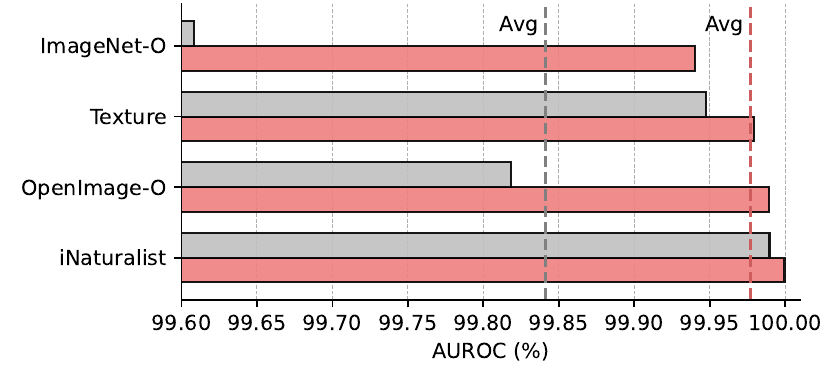}}
    \caption{\revise{The AUROC (\%) of MOODv2 and MOODv1 tested on four OOD datasets, including OpenImage-O \cite{openimages_o}, Texture \cite{dtd}, iNaturalist \cite{inaturalist}, and ImageNet-O \cite{imagenet_o}. }}
  \label{fig:moodv1}
\end{figure}

\begin{figure*}[t]
  \centering
    \includegraphics[width=0.99\linewidth, trim=20 10 20 10]{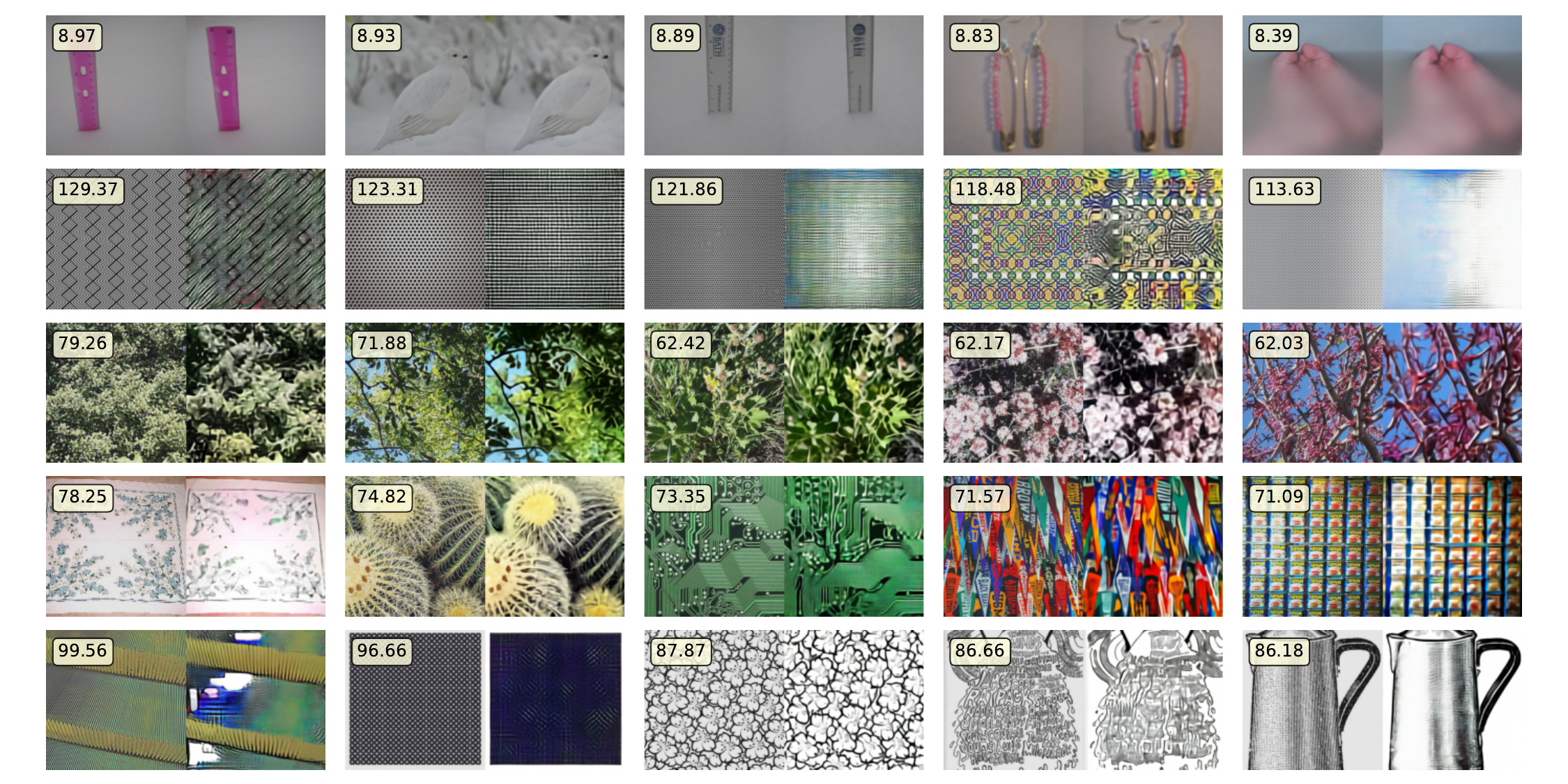}
    \caption{\revise{Each image pair consists of the original image (left) and reconstructed image (right). The rows of images are sourced from ImageNet \cite{imagenet}, Texture \cite{cimpoi14describing}, iNaturalist \cite{van2018inaturalist}, ImageNet-O \cite{hendrycks2021natural}, and OpenImage-O \cite{openimages_o}. The number in the top left corner of each image pair represents the Euclidean distance between the two images.}}
  \label{fig:recover}
\end{figure*}

\begin{table}[t]
\small
\centering

\subfloat[ID: CIFAR-10]{
\setlength{\tabcolsep}{1.1mm}
\begin{tabular}{c|ccccccccc}
\toprule
Methods & prob & feat & logit & feat+logit \\
\midrule
ViT\cite{vit} & \revise{73.61\scriptsize{$\pm$21.36}} & \revise{82.61\scriptsize{$\pm$23.81}} & \revise{45.11\scriptsize{$\pm$4.45}} & \revise{\textbf{99.63}} \\
MoCov3\cite{mocov3} & \revise{70.96\scriptsize{$\pm$23.68}} & \revise{79.17\scriptsize{$\pm$28.75}} & \revise{41.42\scriptsize{$\pm$3.50}} & \revise{\textbf{99.73}} \\
DINOv2\cite{dinov2} & \revise{87.20\scriptsize{$\pm$10.62}} & \revise{84.73\scriptsize{$\pm$21.57}} & \revise{80.30\scriptsize{$\pm$0.10}} & \revise{\textbf{99.98}} \\
\rowcolor{gray!20}BEiTv2\cite{beitv2} & \revise{79.96\scriptsize{$\pm$13.71}} & \revise{91.77\scriptsize{$\pm$11.47}} & \revise{72.87\scriptsize{$\pm$2.08}} & \revise{\textbf{99.87}} \\
\rowcolor{gray!20}BEiT\cite{beit} & \revise{77.51\scriptsize{$\pm$17.83}} & \revise{89.05\scriptsize{$\pm$15.46}} & \revise{65.05\scriptsize{$\pm$2.06}} & \revise{\textbf{99.98}} \\
\bottomrule
\end{tabular}
}

\subfloat[ID: ImageNet]{
\setlength{\tabcolsep}{1.3mm}
\begin{tabular}{c|ccccccccc}
\toprule
Methods & prob & feat & logit & feat+logit \\
\midrule
ViT\cite{vit} & \revise{\textbf{78.52}\scriptsize{$\pm$1.76}} & \revise{76.86\scriptsize{$\pm$3.20}} & \revise{70.61\scriptsize{$\pm$4.76}} & \revise{77.65} \\
MoCov3\cite{mocov3} & \revise{\textbf{78.36}\scriptsize{$\pm$1.42}} & \revise{72.51\scriptsize{$\pm$6.21}} & \revise{70.61\scriptsize{$\pm$5.04}} & \revise{72.07} \\
DINOv2\cite{dinov2} & \revise{59.64\scriptsize{$\pm$7.82}} & \revise{\textbf{63.56}\scriptsize{$\pm$2.89}} & \revise{60.70\scriptsize{$\pm$4.51}} & \revise{61.32} \\
\rowcolor{gray!20}BEiTv2\cite{beitv2} & \revise{89.07\scriptsize{$\pm$0.24}} & \revise{92.96\scriptsize{$\pm$1.27}} & \revise{90.29\scriptsize{$\pm$0.13}} & \revise{\textbf{95.42}} \\
\rowcolor{gray!20}BEiT\cite{beit} & \revise{89.47\scriptsize{$\pm$0.47}} & \revise{93.30\scriptsize{$\pm$1.89}} & \revise{89.84\scriptsize{$\pm$0.01}} & \revise{\textbf{95.68}} \\
\bottomrule
\end{tabular}
}

\caption{\revise{The AUROC (\%) of four types of methods: probability-based methods MSP \cite{baseline_ood} and KL-Matching \cite{maxlogit}; logits-based methods Energy \cite{energy} and MaxLogit \cite{maxlogit}; features-based methods Residual \cite{vim}, React \cite{react} and Mahalanobis \cite{mahalanobis}; and methods using both logits and features include ViM \cite{vim}. The best method for each model is emphasized in bold.}
}
\label{tab:ablation-statistic}
\end{table}

\subsection{MOODv1}
\revise{Our previous version MOODv1 \cite{MOOD} has introduced masked image modeling pretraining strategy into the OOD detection (MOOD) and achieved promising results. However, there are still concerns:}

\revise{Firstly, previous studies \cite{MOOD, csi, ssd} have typically necessitated fine-tuning a model on each in-distribution dataset. The expense of training becomes notably high when dealing with a substantial number of ID datasets to be assessed, such as in one-class OOD detection \cite{csi, MOOD}. However, through experimental validation, we have discovered that a well-prepared masked image modeling model doesn't require additional fine-tuning to achieve outstanding detection performance, conserving substantial fine-tuning resource consumption when dealing with a plethora of ID datasets that require evaluation.}

\revise{Secondly, as the field has seen the emergence of more advanced OOD score functions \cite{maxlogit, vim, react, baseline_ood, energy, mahalanobis} and pretraining techniques \cite{beitv2, dinov2, mocov3, beit, vit}, it raises the question of whether masked image modeling continues to maintain its leading role. In MOODv2, we integrate the latest advancements in pretraining methods and conduct experiments with an array of state-of-the-art OOD score functions. This broader spectrum of pretraining methods and score functions allows for a more comprehensive assessment of the MOODv2's performance, better aligning MOODv2 with the increasingly intricate challenges of OOD detection. }

\revise{Lastly, it is well known that if the network has seen similar samples in training, regardless of pre-training or fine-tuning, the OOD performance will be more or less trivial \cite{openimages_o}. Previous works \cite{oodlimits, MOOD} rely on pre-training on ImageNet-21K, so that the benchmark OOD dataset such as CIFAR \cite{cifar}, Places \cite{places}, etc., is unlikely to be untouched by the ImageNet-21K \cite{imagenet} dataset. In this work, MOODv2 introduces the latest unnatural datasets as OOD, which rules out the possibility of overlap between the OOD test set and the training set \cite{openimages_o, imagenet_o}.}

\revise{In summary, MOODv2 incorporates improved score functions, advanced pretraining techniques, a wider range of unnatural OOD datasets, and a streamlined general framework. The performance improvement of MOODv2 compared to MOODv1 is depicted in Fig. \ref{fig:moodv1}. On ImageNet, MOODv2 exhibits a noteworthy 2.17\% improvement in AUROC compared to MOODv1. Furthermore, on CIFAR-10, MOODv2, without finetuning on the ID dataset, achieves an exceptional AUROC score of up to 99.98\%. }

\section{Methods}\label{sec:methods}

% \input{tables/difference}

% We first define the notations. For a given dataset $X_{\rm ID}$, the goal of out-of-distribution (OOD) detection is to model a detector that identifies whether an input image $x \in X_{\rm ID}$ or $x \notin X_{\rm ID}$ (that is, $x \in X_{\rm OOD}$). A majority of existing methods for OOD detection define an OOD score function $s(x)$. Its abnormal high or low value represents that $x$ is from out-of-distribution. 

\revise{In this section, we initiate our exploration of reconstruction tasks for OOD detection by presenting the underlying motivation in \cref{sec:motivation}. Following that, in \cref{sec:pretask}, we delve into a comprehensive analysis of the essential attributes that play a pivotal role in OOD detection.}

\subsection{Motivation: seeking for effective ID representation} 
\label{sec:motivation}

% In this section, we choose the pretext task that can provide the intrinsic prior to suit the OOD detection task. 
Most previous OOD methods learn the ID representation through classification \cite{baseline_ood, oodlimits} or contrastive learning \cite{csi, ssd} on ID samples, which take advantage of either the ground truth or pseudo labels to supervise the classification networks. On the other hand, work of \cite{backdoor_attack, frog_attack} shows that classification networks only learn different patterns among training categories because it is a shortcut to fulfill classification. \revise{It is indicated that the network actually does not learn the effective in-distribution representation. In comparison, the reconstruction-based pretext task forces the network to learn the pixel-level image representation of the ID images during training to reconstruct the image instead of the patterns for classification. In this way, the network can learn a more representative feature of the ID dataset.}

\iffalse
\begin{equation}
x_{r} = f_{e}\cdot f_{d}(x),
\end{equation}
where $x$ and $x_{r}$ represent the original and reconstructed images. $f_{e}$ and $f_{d}$ correspond to the image encoder and decoder, and $f_{e}$ needs to learn representative features from the training dataset to reconstruct images. To assess the benefits of using the pretask features, we compute the recovery distance between the original and reconstructed images as follows:
\begin{equation}
d_r = \left\lVert x - x_{r} \right\rVert.
\end{equation}
\fi
\revise{To verify this, we reconstruct ID and OOD data and compute the Euclidean distance between the original and reconstructed images. A greater distance indicates a larger deviation of the reconstructed image from the original image. We collect recovery distances for ID and OOD data. Examples of the reconstruction are depicted in \cref{fig:recover}. In the first row, for ID images, pre-trained models reconstruct the images effectively. Instead, for unnatural OOD images in the following rows, clear domain discrepancies emerge. For instance, in the case of textured images, the models still apply lighting and shadows reminiscent of natural images. In the case of sketch images,  the models render the images smoother and brighter. This discrepancy visually highlights the domain gap in model features between ID and OOD data, which can be leveraged for OOD detection.}

\begin{figure*}[t]
  \centering
    \subfloat[ID: CIFAR-10]{\includegraphics[width=0.99\linewidth, trim=10 15 5 15]{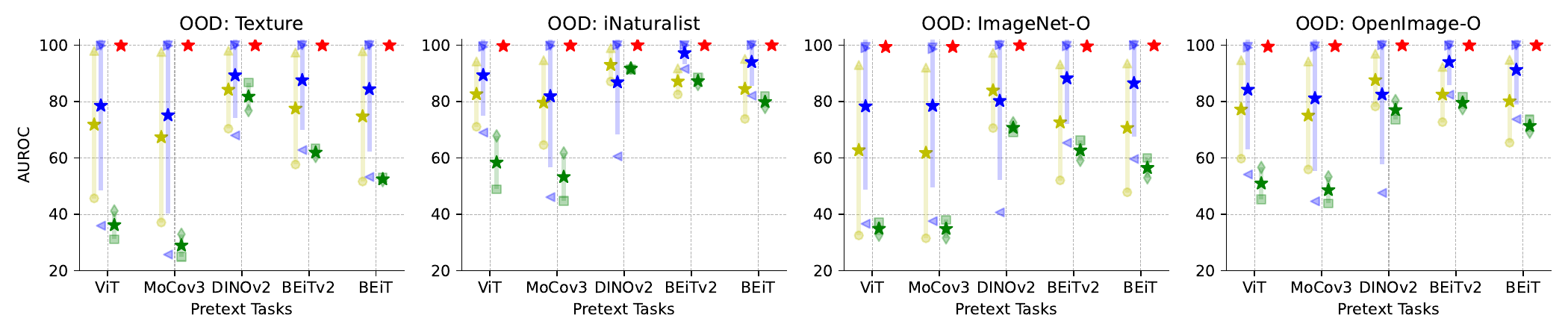}}
    
    \subfloat[ID: ImageNet]{\includegraphics[width=0.99\linewidth, trim=10 15 5 15]{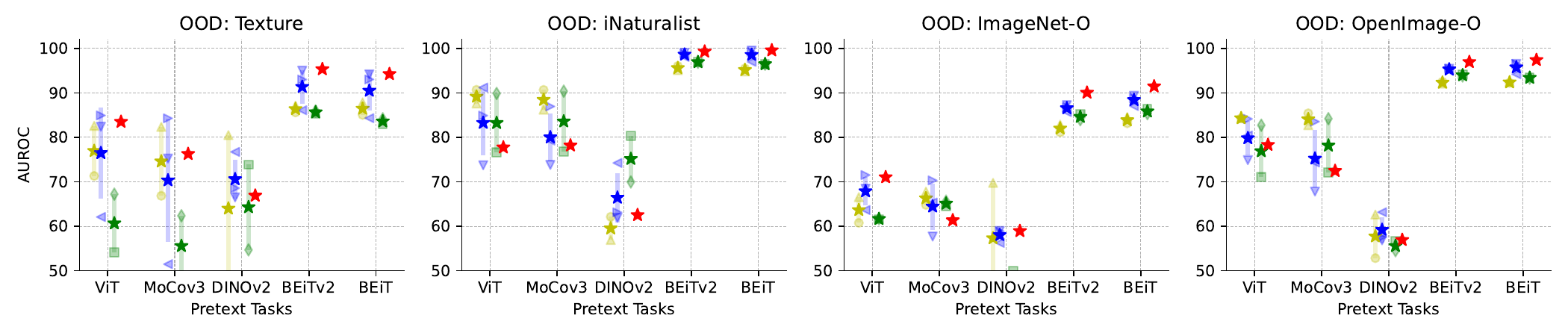}}

    \caption{\revise{The AUROC (\%) tested on unnatural OOD datasets of various OOD detection algorithms applied to a ViT model. 
    The pre-text tasks include classification task \cite{vit}, contrastive learning tasks MoCov3 \cite{mocov3} and DINOv2 \cite{dinov2}, and masked image modeling tasks BEiT series \cite{beit,beitv2}.
    Methods in blue utilize the feature space;
    methods in green use logits;
    methods in yellow make use of the softmax probability.
    and methods in red leverage both features and logits.
    Stars represent the average AUROC for methods in the corresponding colors; light vertical lines represent the standard deviation.
    }}
    
    % probability-based methods MSP \cite{baseline_ood} and KL-Matching \cite{maxlogit}; logits-based methods Energy \cite{energy} and MaxLogit \cite{maxlogit}; features-based methods Residual \cite{vim}, React \cite{react} and Mahalanobis \cite{mahalanobis}; and methods using both logits and features include ViM \cite{vim}.
    
  \label{fig:detailed_ablation}
\end{figure*}

\subsection{Reconstruction Tasks for OOD Detection} \label{sec:pretask}
%--------------------------------------multi-class-----------------------------------
\begin{table*}[tp]
\small
\centering
\setlength{\tabcolsep}{0.6mm}
\begin{tabular}{c|c|ccccccccccc}
\toprule
\multirow{2}{*}{Methods} & \multirow{2}{*}{Models} & \multicolumn{2}{c}{Texture \cite{dtd}} & \multicolumn{2}{c}{iNaturalist \cite{inaturalist}} & \multicolumn{2}{c}{ImageNet-O \cite{imagenet_o}} & \multicolumn{2}{c}{ OpenImage-O \cite{openimages_o}} & \multicolumn{2}{c}{Average} \\
& & AUROC$\uparrow$ & FPR95$\downarrow$ & AUROC$\uparrow$ & FPR95$\downarrow$ & AUROC$\uparrow$ & FPR95$\downarrow$ & AUROC$\uparrow$ & FPR95$\downarrow$ & AUROC$\uparrow$ & FPR95$\downarrow$ \\
\midrule
\multirow{5}{*}{MSP\cite{baseline_ood}}
& ViT\cite{vit} & \revise{71.31} & \revise{71.31} & \revise{90.70} & \revise{90.70} & \revise{60.77} & \revise{60.77} & \revise{84.29} & \revise{84.29} & \revise{76.77} & \revise{76.77} \\
& MoCov3\cite{mocov3} & \revise{66.85} & \revise{66.85} & \revise{90.68} & \revise{90.68} & \revise{64.80} & \revise{64.80} & \revise{85.42} & \revise{85.42} & \revise{76.94} & \revise{76.94} \\
& DINOv2\cite{dinov2} & \revise{47.49} & \revise{47.49} & \revise{62.13} & \revise{62.13} & \revise{44.87} & \revise{44.87} & \revise{52.83} & \revise{52.83} & \revise{51.83} & \revise{51.83} \\
& \cellcolor{gray!20}{BEiTv2\cite{beitv2}} & \cellcolor{gray!20}{\revise{\textbf{85.61}}} & \cellcolor{gray!20}{\revise{\textbf{85.61}}} & \cellcolor{gray!20}{\revise{\textbf{96.05}}} & \cellcolor{gray!20}{\revise{\textbf{96.05}}} & \cellcolor{gray!20}{\revise{81.15}} & \cellcolor{gray!20}{\revise{81.15}} & \cellcolor{gray!20}{\revise{\textbf{92.52}}} & \cellcolor{gray!20}{\revise{\textbf{92.52}}} & \cellcolor{gray!20}{\revise{88.83}} & \cellcolor{gray!20}{\revise{88.83}} \\
& \cellcolor{gray!20}{BEiT\cite{beit}} & \cellcolor{gray!20}{\revise{85.05}} & \cellcolor{gray!20}{\revise{85.05}} & \cellcolor{gray!20}{\revise{95.50}} & \cellcolor{gray!20}{\revise{95.50}} & \cellcolor{gray!20}{\revise{\textbf{83.17}}} & \cellcolor{gray!20}{\revise{\textbf{83.17}}} & \cellcolor{gray!20}{\revise{92.28}} & \cellcolor{gray!20}{\revise{92.28}} & \cellcolor{gray!20}{\revise{\textbf{89.00}}} & \cellcolor{gray!20}{\revise{\textbf{89.00}}} \\
\midrule
\multirow{5}{*}{Energy\cite{energy}}
& ViT\cite{vit} & \revise{54.11} & \revise{54.11} & \revise{76.61} & \revise{76.61} & \revise{61.63} & \revise{61.63} & \revise{71.06} & \revise{71.06} & \revise{65.85} & \revise{65.85} \\
& MoCov3\cite{mocov3} & \revise{48.79} & \revise{48.79} & \revise{76.80} & \revise{76.80} & \revise{64.56} & \revise{64.56} & \revise{72.13} & \revise{72.13} & \revise{65.57} & \revise{65.57} \\
& DINOv2\cite{dinov2} & \revise{73.89} & \revise{73.89} & \revise{80.34} & \revise{80.34} & \revise{49.98} & \revise{49.98} & \revise{56.64} & \revise{56.64} & \revise{65.21} & \revise{65.21} \\
& \cellcolor{gray!20}{BEiTv2\cite{beitv2}} & \cellcolor{gray!20}{\revise{\textbf{85.32}}} & \cellcolor{gray!20}{\revise{\textbf{85.32}}} & \cellcolor{gray!20}{\revise{\textbf{96.95}}} & \cellcolor{gray!20}{\revise{\textbf{96.95}}} & \cellcolor{gray!20}{\revise{85.27}} & \cellcolor{gray!20}{\revise{85.27}} & \cellcolor{gray!20}{\revise{\textbf{94.14}}} & \cellcolor{gray!20}{\revise{\textbf{94.14}}} & \cellcolor{gray!20}{\revise{\textbf{90.42}}} & \cellcolor{gray!20}{\revise{\textbf{90.42}}} \\
& \cellcolor{gray!20}{BEiT\cite{beit}} & \cellcolor{gray!20}{\revise{83.04}} & \cellcolor{gray!20}{\revise{83.04}} & \cellcolor{gray!20}{\revise{96.48}} & \cellcolor{gray!20}{\revise{96.48}} & \cellcolor{gray!20}{\revise{\textbf{86.36}}} & \cellcolor{gray!20}{\revise{\textbf{86.36}}} & \cellcolor{gray!20}{\revise{93.50}} & \cellcolor{gray!20}{\revise{93.50}} & \cellcolor{gray!20}{\revise{89.85}} & \cellcolor{gray!20}{\revise{89.85}} \\
\midrule
\multirow{5}{*}{MaxLogit\cite{maxlogit}}
& ViT\cite{vit} & \revise{67.22} & \revise{67.22} & \revise{89.88} & \revise{89.88} & \revise{61.68} & \revise{61.68} & \revise{82.73} & \revise{82.73} & \revise{75.37} & \revise{75.37} \\
& MoCov3\cite{mocov3} & \revise{62.36} & \revise{62.36} & \revise{90.38} & \revise{90.38} & \revise{65.65} & \revise{65.65} & \revise{84.19} & \revise{84.19} & \revise{75.64} & \revise{75.64} \\
& DINOv2\cite{dinov2} & \revise{54.70} & \revise{54.70} & \revise{69.98} & \revise{69.98} & \revise{45.60} & \revise{45.60} & \revise{54.52} & \revise{54.52} & \revise{56.20} & \revise{56.20} \\
& \cellcolor{gray!20}{BEiTv2\cite{beitv2}} & \cellcolor{gray!20}{\revise{\textbf{85.94}}} & \cellcolor{gray!20}{\revise{\textbf{85.94}}} & \cellcolor{gray!20}{\revise{\textbf{96.90}}} & \cellcolor{gray!20}{\revise{\textbf{96.90}}} & \cellcolor{gray!20}{\revise{83.97}} & \cellcolor{gray!20}{\revise{83.97}} & \cellcolor{gray!20}{\revise{\textbf{93.82}}} & \cellcolor{gray!20}{\revise{\textbf{93.82}}} & \cellcolor{gray!20}{\revise{\textbf{90.16}}} & \cellcolor{gray!20}{\revise{\textbf{90.16}}} \\
& \cellcolor{gray!20}{BEiT\cite{beit}} & \cellcolor{gray!20}{\revise{84.17}} & \cellcolor{gray!20}{\revise{84.17}} & \cellcolor{gray!20}{\revise{96.48}} & \cellcolor{gray!20}{\revise{96.48}} & \cellcolor{gray!20}{\revise{\textbf{85.34}}} & \cellcolor{gray!20}{\revise{\textbf{85.34}}} & \cellcolor{gray!20}{\revise{93.31}} & \cellcolor{gray!20}{\revise{93.31}} & \cellcolor{gray!20}{\revise{89.83}} & \cellcolor{gray!20}{\revise{89.83}} \\
\midrule
\multirow{5}{*}{KL-Matching\cite{maxlogit}}
& ViT\cite{vit} & \revise{82.59} & \revise{82.59} & \revise{87.63} & \revise{87.63} & \revise{66.55} & \revise{66.55} & \revise{84.34} & \revise{84.34} & \revise{80.28} & \revise{80.28} \\
& MoCov3\cite{mocov3} & \revise{82.35} & \revise{82.35} & \revise{86.24} & \revise{86.24} & \revise{67.80} & \revise{67.80} & \revise{82.73} & \revise{82.73} & \revise{79.78} & \revise{79.78} \\
& DINOv2\cite{dinov2} & \revise{80.51} & \revise{80.51} & \revise{56.93} & \revise{56.93} & \revise{69.77} & \revise{69.77} & \revise{62.63} & \revise{62.63} & \revise{67.46} & \revise{67.46} \\
& \cellcolor{gray!20}{BEiTv2\cite{beitv2}} & \cellcolor{gray!20}{\revise{87.14}} & \cellcolor{gray!20}{\revise{87.14}} & \cellcolor{gray!20}{\revise{\textbf{95.13}}} & \cellcolor{gray!20}{\revise{\textbf{95.13}}} & \cellcolor{gray!20}{\revise{82.87}} & \cellcolor{gray!20}{\revise{82.87}} & \cellcolor{gray!20}{\revise{92.10}} & \cellcolor{gray!20}{\revise{92.10}} & \cellcolor{gray!20}{\revise{89.31}} & \cellcolor{gray!20}{\revise{89.31}} \\
& \cellcolor{gray!20}{BEiT\cite{beit}} & \cellcolor{gray!20}{\revise{\textbf{87.87}}} & \cellcolor{gray!20}{\revise{\textbf{87.87}}} & \cellcolor{gray!20}{\revise{94.82}} & \cellcolor{gray!20}{\revise{94.82}} & \cellcolor{gray!20}{\revise{\textbf{84.56}}} & \cellcolor{gray!20}{\revise{\textbf{84.56}}} & \cellcolor{gray!20}{\revise{\textbf{92.48}}} & \cellcolor{gray!20}{\revise{\textbf{92.48}}} & \cellcolor{gray!20}{\revise{\textbf{89.93}}} & \cellcolor{gray!20}{\revise{\textbf{89.93}}} \\
\midrule
\multirow{5}{*}{Residual\cite{vim}}
& ViT\cite{vit} & \revise{82.39} & \revise{82.39} & \revise{73.72} & \revise{73.72} & \revise{68.44} & \revise{68.44} & \revise{74.88} & \revise{74.88} & \revise{74.86} & \revise{74.86} \\
& MoCov3\cite{mocov3} & \revise{75.25} & \revise{75.25} & \revise{73.80} & \revise{73.80} & \revise{57.69} & \revise{57.69} & \revise{67.82} & \revise{67.82} & \revise{68.64} & \revise{68.64} \\
& DINOv2\cite{dinov2} & \revise{66.50} & \revise{66.50} & \revise{61.90} & \revise{61.90} & \revise{58.94} & \revise{58.94} & \revise{56.84} & \revise{56.84} & \revise{61.04} & \revise{61.04} \\
& \cellcolor{gray!20}{BEiTv2\cite{beitv2}} & \cellcolor{gray!20}{\revise{\textbf{94.99}}} & \cellcolor{gray!20}{\revise{\textbf{94.99}}} & \cellcolor{gray!20}{\revise{99.01}} & \cellcolor{gray!20}{\revise{99.01}} & \cellcolor{gray!20}{\revise{87.23}} & \cellcolor{gray!20}{\revise{87.23}} & \cellcolor{gray!20}{\revise{95.43}} & \cellcolor{gray!20}{\revise{95.43}} & \cellcolor{gray!20}{\revise{94.17}} & \cellcolor{gray!20}{\revise{94.17}} \\
& \cellcolor{gray!20}{BEiT\cite{beit}} & \cellcolor{gray!20}{\revise{94.16}} & \cellcolor{gray!20}{\revise{94.16}} & \cellcolor{gray!20}{\revise{\textbf{99.50}}} & \cellcolor{gray!20}{\revise{\textbf{99.50}}} & \cellcolor{gray!20}{\revise{\textbf{89.35}}} & \cellcolor{gray!20}{\revise{\textbf{89.35}}} & \cellcolor{gray!20}{\revise{\textbf{96.52}}} & \cellcolor{gray!20}{\revise{\textbf{96.52}}} & \cellcolor{gray!20}{\revise{\textbf{94.88}}} & \cellcolor{gray!20}{\revise{\textbf{94.88}}} \\
\midrule
\multirow{5}{*}{React\cite{react}}
& ViT\cite{vit} & \revise{62.09} & \revise{62.09} & \revise{91.20} & \revise{91.20} & \revise{63.66} & \revise{63.66} & \revise{80.43} & \revise{80.43} & \revise{74.34} & \revise{74.34} \\
& MoCov3\cite{mocov3} & \revise{51.47} & \revise{51.47} & \revise{79.30} & \revise{79.30} & \revise{65.33} & \revise{65.33} & \revise{74.35} & \revise{74.35} & \revise{67.61} & \revise{67.61} \\
& DINOv2\cite{dinov2} & \revise{76.73} & \revise{76.73} & \revise{74.25} & \revise{74.25} & \revise{56.26} & \revise{56.26} & \revise{63.17} & \revise{63.17} & \revise{67.60} & \revise{67.60} \\
& \cellcolor{gray!20}{BEiTv2\cite{beitv2}} & \cellcolor{gray!20}{\revise{\textbf{86.10}}} & \cellcolor{gray!20}{\revise{\textbf{86.10}}} & \cellcolor{gray!20}{\revise{\textbf{98.09}}} & \cellcolor{gray!20}{\revise{\textbf{98.09}}} & \cellcolor{gray!20}{\revise{85.69}} & \cellcolor{gray!20}{\revise{85.69}} & \cellcolor{gray!20}{\revise{\textbf{94.96}}} & \cellcolor{gray!20}{\revise{\textbf{94.96}}} & \cellcolor{gray!20}{\revise{\textbf{91.21}}} & \cellcolor{gray!20}{\revise{\textbf{91.21}}} \\
& \cellcolor{gray!20}{BEiT\cite{beit}} & \cellcolor{gray!20}{\revise{84.32}} & \cellcolor{gray!20}{\revise{84.32}} & \cellcolor{gray!20}{\revise{96.99}} & \cellcolor{gray!20}{\revise{96.99}} & \cellcolor{gray!20}{\revise{\textbf{87.04}}} & \cellcolor{gray!20}{\revise{\textbf{87.04}}} & \cellcolor{gray!20}{\revise{94.21}} & \cellcolor{gray!20}{\revise{94.21}} & \cellcolor{gray!20}{\revise{90.64}} & \cellcolor{gray!20}{\revise{90.64}} \\
\midrule
\multirow{5}{*}{Mahalanobis\cite{mahalanobis}}
& ViT\cite{vit} & \revise{84.93} & \revise{84.93} & \revise{84.90} & \revise{84.90} & \revise{71.53} & \revise{71.53} & \revise{84.16} & \revise{84.16} & \revise{81.38} & \revise{81.38} \\
& MoCov3\cite{mocov3} & \revise{84.29} & \revise{84.29} & \revise{86.95} & \revise{86.95} & \revise{70.33} & \revise{70.33} & \revise{83.54} & \revise{83.54} & \revise{81.28} & \revise{81.28} \\
& DINOv2\cite{dinov2} & \revise{68.58} & \revise{68.58} & \revise{63.14} & \revise{63.14} & \revise{58.86} & \revise{58.86} & \revise{57.57} & \revise{57.57} & \revise{62.04} & \revise{62.04} \\
& \cellcolor{gray!20}{BEiTv2\cite{beitv2}} & \cellcolor{gray!20}{\revise{93.01}} & \cellcolor{gray!20}{\revise{93.01}} & \cellcolor{gray!20}{\revise{98.78}} & \cellcolor{gray!20}{\revise{98.78}} & \cellcolor{gray!20}{\revise{86.78}} & \cellcolor{gray!20}{\revise{86.78}} & \cellcolor{gray!20}{\revise{95.46}} & \cellcolor{gray!20}{\revise{95.46}} & \cellcolor{gray!20}{\revise{93.51}} & \cellcolor{gray!20}{\revise{93.51}} \\
& \cellcolor{gray!20}{BEiT\cite{beit}} & \cellcolor{gray!20}{\revise{\textbf{93.03}}} & \cellcolor{gray!20}{\revise{\textbf{93.03}}} & \cellcolor{gray!20}{\revise{\textbf{99.18}}} & \cellcolor{gray!20}{\revise{\textbf{99.18}}} & \cellcolor{gray!20}{\revise{\textbf{88.84}}} & \cellcolor{gray!20}{\revise{\textbf{88.84}}} & \cellcolor{gray!20}{\revise{\textbf{96.51}}} & \cellcolor{gray!20}{\revise{\textbf{96.51}}} & \cellcolor{gray!20}{\revise{\textbf{94.39}}} & \cellcolor{gray!20}{\revise{\textbf{94.39}}} \\
\midrule
\multirow{5}{*}{ViM\cite{vim}}
& ViT\cite{vit} & \revise{83.51} & \revise{83.51} & \revise{77.75} & \revise{77.75} & \revise{71.04} & \revise{71.04} & \revise{78.31} & \revise{78.31} & \revise{77.65} & \revise{77.65} \\
& MoCov3\cite{mocov3} & \revise{76.28} & \revise{76.28} & \revise{78.18} & \revise{78.18} & \revise{61.35} & \revise{61.35} & \revise{72.46} & \revise{72.46} & \revise{72.07} & \revise{72.07} \\
& DINOv2\cite{dinov2} & \revise{66.90} & \revise{66.90} & \revise{62.53} & \revise{62.53} & \revise{58.93} & \revise{58.93} & \revise{56.93} & \revise{56.93} & \revise{61.32} & \revise{61.32} \\
& \cellcolor{gray!20}{BEiTv2\cite{beitv2}} & \cellcolor{gray!20}{\revise{\textbf{95.35}}} & \cellcolor{gray!20}{\revise{\textbf{95.35}}} & \cellcolor{gray!20}{\revise{99.31}} & \cellcolor{gray!20}{\revise{99.31}} & \cellcolor{gray!20}{\revise{90.06}} & \cellcolor{gray!20}{\revise{90.06}} & \cellcolor{gray!20}{\revise{96.96}} & \cellcolor{gray!20}{\revise{96.96}} & \cellcolor{gray!20}{\revise{95.42}} & \cellcolor{gray!20}{\revise{95.42}} \\
& \cellcolor{gray!20}{BEiT\cite{beit}} & \cellcolor{gray!20}{\revise{94.25}} & \cellcolor{gray!20}{\revise{94.25}} & \cellcolor{gray!20}{\revise{\textbf{99.59}}} & \cellcolor{gray!20}{\revise{\textbf{99.59}}} & \cellcolor{gray!20}{\revise{\textbf{91.47}}} & \cellcolor{gray!20}{\revise{\textbf{91.47}}} & \cellcolor{gray!20}{\revise{\textbf{97.41}}} & \cellcolor{gray!20}{\revise{\textbf{97.41}}} & \cellcolor{gray!20}{\revise{\textbf{95.68}}} & \cellcolor{gray!20}{\revise{\textbf{95.68}}} \\
\midrule
\multirow{5}{*}{Average}
& ViT\cite{vit} & \revise{73.52} & \revise{73.52} & \revise{84.05} & \revise{84.05} & \revise{65.66} & \revise{65.66} & \revise{80.02} & \revise{80.02} & \revise{75.81} & \revise{75.81} \\
& MoCov3\cite{mocov3} & \revise{68.45} & \revise{68.45} & \revise{82.79} & \revise{82.79} & \revise{64.69} & \revise{64.69} & \revise{77.83} & \revise{77.83} & \revise{73.44} & \revise{73.44} \\
& DINOv2\cite{dinov2} & \revise{66.91} & \revise{66.91} & \revise{66.40} & \revise{66.40} & \revise{55.40} & \revise{55.40} & \revise{57.64} & \revise{57.64} & \revise{61.59} & \revise{61.59} \\
& \cellcolor{gray!20}{BEiTv2\cite{beitv2}} & \cellcolor{gray!20}{\revise{\textbf{89.18}}} & \cellcolor{gray!20}{\revise{\textbf{89.18}}} & \cellcolor{gray!20}{\revise{\textbf{97.53}}} & \cellcolor{gray!20}{\revise{\textbf{97.53}}} & \cellcolor{gray!20}{\revise{85.38}} & \cellcolor{gray!20}{\revise{85.38}} & \cellcolor{gray!20}{\revise{94.42}} & \cellcolor{gray!20}{\revise{94.42}} & \cellcolor{gray!20}{\revise{91.63}} & \cellcolor{gray!20}{\revise{91.63}} \\
& \cellcolor{gray!20}{BEiT\cite{beit}} & \cellcolor{gray!20}{\revise{88.24}} & \cellcolor{gray!20}{\revise{88.24}} & \cellcolor{gray!20}{\revise{97.32}} & \cellcolor{gray!20}{\revise{97.32}} & \cellcolor{gray!20}{\revise{\textbf{87.02}}} & \cellcolor{gray!20}{\revise{\textbf{87.02}}} & \cellcolor{gray!20}{\revise{\textbf{94.53}}} & \cellcolor{gray!20}{\revise{\textbf{94.53}}} & \cellcolor{gray!20}{\revise{\textbf{91.77}}} & \cellcolor{gray!20}{\revise{\textbf{91.77}}} \\
\midrule
\multirow{5}{*}{Best}
& ViT\cite{vit} & \revise{84.93} & \revise{84.93} & \revise{91.20} & \revise{91.20} & \revise{71.53} & \revise{71.53} & \revise{84.34} & \revise{84.34} & \revise{81.38} & \revise{81.38} \\
& MoCov3\cite{mocov3} & \revise{84.29} & \revise{84.29} & \revise{90.68} & \revise{90.68} & \revise{70.33} & \revise{70.33} & \revise{85.42} & \revise{85.42} & \revise{81.28} & \revise{81.28} \\
& DINOv2\cite{dinov2} & \revise{80.51} & \revise{80.51} & \revise{80.34} & \revise{80.34} & \revise{69.77} & \revise{69.77} & \revise{63.17} & \revise{63.17} & \revise{67.60} & \revise{67.60} \\
& \cellcolor{gray!20}{BEiTv2\cite{beitv2}} & \cellcolor{gray!20}{\revise{\textbf{95.35}}} & \cellcolor{gray!20}{\revise{\textbf{95.35}}} & \cellcolor{gray!20}{\revise{99.31}} & \cellcolor{gray!20}{\revise{99.31}} & \cellcolor{gray!20}{\revise{90.06}} & \cellcolor{gray!20}{\revise{90.06}} & \cellcolor{gray!20}{\revise{96.96}} & \cellcolor{gray!20}{\revise{96.96}} & \cellcolor{gray!20}{\revise{95.42}} & \cellcolor{gray!20}{\revise{95.42}} \\
& \cellcolor{gray!20}{BEiT\cite{beit}} & \cellcolor{gray!20}{\revise{94.25}} & \cellcolor{gray!20}{\revise{94.25}} & \cellcolor{gray!20}{\revise{\textbf{99.59}}} & \cellcolor{gray!20}{\revise{\textbf{99.59}}} & \cellcolor{gray!20}{\revise{\textbf{91.47}}} & \cellcolor{gray!20}{\revise{\textbf{91.47}}} & \cellcolor{gray!20}{\revise{\textbf{97.41}}} & \cellcolor{gray!20}{\revise{\textbf{97.41}}} & \cellcolor{gray!20}{\revise{\textbf{95.68}}} & \cellcolor{gray!20}{\revise{\textbf{95.68}}} \\
\bottomrule
\end{tabular}
\caption{
\revise{Performance of OOD detection methods on ViT-B/16 model with $224\times224$-pixel inputs. The pre-text tasks include classification task \cite{vit}, contrastive learning tasks MoCov3 \cite{mocov3} and DINOv2 \cite{dinov2}, and masked image modeling tasks BEiT \cite{beit} and BEiTv2 \cite{beitv2}. All models are per-trained on ImageNet-21k and finetuned on ImageNet-1k. Both metrics AUROC and FPR95 are in percentage.
The best method is emphasized in bold and a gray background indicates our choice.}
}
\label{tab:multi-class-imagenet-ablation}
\end{table*}

\revise{In this section, we offer a comprehensive analysis of these key elements in the context of OOD detection. We employ ImageNet \cite{imagenet} as the in-distribution dataset and evaluate pre-task texts on challenging unnatural out-of-distribution datasets, including OpenImage-O \cite{openimages_o}, Texture \cite{dtd}, iNaturalist \cite{inaturalist}, and ImageNet-O \cite{imagenet_o}. Extensive validations with various pretraining methods and OOD score functions, including {MSP} \cite{baseline_ood}, {Energy} \cite{energy}, {ODIN} \cite{odin}, {MaxLogit} \cite{maxlogit}, {KL Matching} \cite{maxlogit}, {Residual} \cite{vim}, {ReAct} \cite{react}, {Mahalanobis} \cite{mahalanobis} and {ViM} \cite{vim}. }

\revise{Results are shown in ~\cref{tab:multi-class-imagenet-ablation}. The results indicate that the masked image modeling pretext task surpasses classification and contrastive learning pretext tasks when employing all included score functions. The average AUROC across these score functions exhibits an improvement of 15.96\%  compared to the competition. Models when using the best-performing score function saw a 14.30\% increase in performance. This remarkable achievement can be attributed to the representative ID feature space representation, thereby aiding in distinguishing between ID and OOD data. This discovery is highly significant as it enhances performance across mainstream OOD detection score functions, thus advancing the entire field. We also employ CIFAR-10 \cite{cifar} as the ID dataset and provide results in the appendix. Our approach attains an impressive AUROC of 99.99\% while concurrently reducing the FPR95 to a mere 0.03\%.}

\revise{To enhance the comprehensibility of our experimental findings, we conduct a thorough statistical analysis and illustrate them in visual representations. The outcomes are depicted in \cref{fig:detailed_ablation}. Our approach not only leads to an overall enhancement in results but also notably minimizes the variations among different methods. For instance, the ViT, MoCov3, and DINOv2 models using logit-based methods exhibited standard deviations of 4.76\%, 5.04\%, and 4.51\%, respectively, while BEiT and BEiTv2 displayed significantly lower standard deviations, reaching as low as 0.13\% and 0.01\%. This observation underscores that even uncomplicated score functions can perform equivalently to more intricate ones when an effective ID feature representation is applied.}

\revise{In \cref{tab:ablation-statistic}, we underscore the optimal methods for each model. On CIFAR-10, all models achieved their best results when employing the feat and logit combination approach, achieving almost 100\% accuracy. This suggests a highly effective grasp of CIFAR-10's feature space. Conversely, with the larger ImageNet dataset, we observed variations in outcomes. Notably, the masked image modeling pretext-pretrained model achieved the best results when using the feat and logit combination method, while other models excelled in probability-based and feature-based methods. Additionally, our masked image modeling pretext demonstrated significantly superior performance compared to other pretraining methods, underscoring the limitations of classification-based pretraining strategies and their inadequacy in harnessing advanced score functions effectively. These discoveries reinforce the pivotal role of proficient feature representation in OOD detection. Furthermore, for more detailed information, we provide illustrations of the distribution curves of OOD scores for both ID and OOD datasets in the appendix.}

\subsection{Masked Image Modeling for Out-of-Distribution v2} 
\label{sec:method}
\revise{To sum up, in this section, we observed that pre-trained models adeptly reconstruct ID images, yet manifest distinctive domain differences in the OOD scenario (\cref{fig:recover}). This visual incongruity starkly highlights the prevailing domain gap in model features between ID and OOD data. Additionally, a thorough analysis of experimental outcomes reveals that the pre-task of masked image modeling not only significantly enhances overall results but also markedly diminishes disparities among score functions. These findings emphasize the crucial significance of effective feature representation in OOD detection, highlighting the enhancement of features through masked image modeling tasks.}

\revise{Finally, we propose our Masked Image Modeling for Out-of-Distribution Detection v2 (MOODv2). The algorithm of is shown in \cref{alg1}, mainly including the following stages.}

\begin{enumerate}
\item \revise{Pre-train the vision encoder with masked image modeling on the pretrain dataset. }
\item \revise{Apply fine-tuning the backbone on the in-distribution dataset.}
\item \revise{Extract features from the trained image encoder and calculate the OOD score distance score function for OOD detection.}
\end{enumerate}

\revise{In terms of the OOD score function, we adopt  ViM\cite{vim} that combines features and logits, leveraging insights from the masked image modeling pre-trained model, which has demonstrated superior performance. Mathematically, the score is
\begin{equation}
    \text{s}(x) = \frac{e^{\alpha\sqrt{x^T R R^Tx}}}{\sum_{i=1}^C e^{l_i} + e^{\alpha\sqrt{x^TRR^Tx}}}.
\end{equation}
where $l_i$ is the $i$-th logit of feature $x$ in the training set $X$; $\alpha$ is a per-model constant; $R\in\mathbb{R}^{N\times(N-D)}$ is the $(D+1)$-th column to the last column of the eigenvector matrix $Q$ of $X$ and $N$ is the principal dimension; $C$ is the number of classes.}

%--------------------------------------algorithm-----------------------------------
\begin{algorithm}[t!]
% \color{blue}
\caption{MOODv2 Detection Algorithm}
\label{alg1}
\small
\begin{algorithmic}[1]

\Require Pre-train set $X_P$, in-distribution set $X_{\rm ID}$, test set $X_{\rm test}$, required True Positive Rate $\eta$\%, backbone $f$.
\Ensure Is $x_{\rm test}$ outlier or not? $\forall x_{\rm test} \in X_{\rm test}$.

\State Pre-train $f$ on $X_P$ by maximizing
$$\sum_{x\in X_P}\mathbb{E}_M\left [\sum_{i\in M}\log p_{\rm MIM}(z|x^M)\right ]$$.
\State Fine-tune $f$ on $X_P$ by minimizing
$$L_{\rm ft}=\sum_{x_p\in X_P}{\rm CrossEntropy}(f(x_p), y_P(x_p))$$ 
\State Calculate $d(x_{\rm test})$ for $x_{\rm test} \in X_{\rm test}$ and $d(x_{\rm cal})$ for $x_{\rm cal}\in X_{\rm cal}$.
\State Compute threshold $T$ as the $\eta$ percentile of $d(x_{\rm cal})$.
\If {$d(x_{\rm test})>T$}
\State $x_{\rm test}$ is an outlier.
\EndIf
\end{algorithmic}
\end{algorithm}

\begin{table*}[t!]
\small
\centering
\setlength{\tabcolsep}{0.8mm}
\begin{tabular}{c|c|ccccccccccc}
\toprule
\multirow{2}{*}{ID data} & \multirow{2}{*}{Methods} & \multicolumn{2}{c}{Texture \cite{dtd}} & \multicolumn{2}{c}{iNaturalist \cite{inaturalist}} & \multicolumn{2}{c}{ImageNet-O \cite{imagenet_o}} & \multicolumn{2}{c}{ OpenImage-O \cite{openimages_o}} & \multicolumn{2}{c}{Average} \\
& & AUROC$\uparrow$ & FPR95$\downarrow$ & AUROC$\uparrow$ & FPR95$\downarrow$ & AUROC$\uparrow$ & FPR95$\downarrow$ & AUROC$\uparrow$ & FPR95$\downarrow$ & AUROC$\uparrow$ & FPR95$\downarrow$ \\
\midrule
\multirow{10}{*}{CIFAR-10}
& MSP\cite{baseline_ood} & \revise{45.67} & \revise{95.17} & \revise{71.07} & \revise{81.76} & \revise{32.52} & \revise{98.85} & \revise{59.74} & \revise{91.45} & \revise{52.25} & \revise{91.81} \\
& Energy\cite{energy} & \revise{31.16} & \revise{97.89} & \revise{48.95} & \revise{97.92} & \revise{37.22} & \revise{97.85} & \revise{45.29} & \revise{96.36} & \revise{40.65} & \revise{97.50} \\
& MaxLogit\cite{maxlogit} & \revise{41.21} & \revise{95.95} & \revise{67.83} & \revise{86.04} & \revise{32.58} & \revise{98.80} & \revise{56.64} & \revise{92.94} & \revise{49.56} & \revise{93.43} \\
& KL-Matching\cite{maxlogit} & \revise{98.00} & \revise{10.64} & \revise{94.23} & \revise{35.86} & \revise{92.99} & \revise{32.40} & \revise{94.68} & \revise{27.92} & \revise{94.97} & \revise{26.71} \\
& Residual\cite{vim} & \revise{99.91} & \revise{0.21} & \revise{99.68} & \revise{0.45} & \revise{99.36} & \revise{2.85} & \revise{99.42} & \revise{2.46} & \revise{99.59} & \revise{1.49} \\
& React\cite{react} & \revise{35.97} & \revise{96.26} & \revise{69.01} & \revise{87.91} & \revise{36.65} & \revise{97.75} & \revise{54.14} & \revise{93.11} & \revise{48.94} & \revise{93.76} \\
& Mahalanobis\cite{mahalanobis} & \revise{99.77} & \revise{0.60} & \revise{99.39} & \revise{1.11} & \revise{98.93} & \revise{4.90} & \revise{99.14} & \revise{3.26} & \revise{99.31} & \revise{2.47} \\
& ViM\cite{vim} & \revise{99.91} & \revise{0.23} & \revise{99.72} & \revise{0.38} & \revise{99.38} & \revise{2.65} & \revise{99.49} & \revise{2.31} & \revise{99.63} & \revise{1.39} \\
& \cellcolor{gray!20}{MOODv1\cite{MOOD}} & \cellcolor{gray!20}{\revise{99.95}} & \cellcolor{gray!20}{\revise{\textbf{0.06}}} & \cellcolor{gray!20}{\revise{99.99}} & \cellcolor{gray!20}{\revise{0.02}} & \cellcolor{gray!20}{\revise{99.61}} & \cellcolor{gray!20}{\revise{1.90}} & \cellcolor{gray!20}{\revise{99.82}} & \cellcolor{gray!20}{\revise{0.77}} & \cellcolor{gray!20}{\revise{99.84}} & \cellcolor{gray!20}{\revise{0.69}} \\
& \cellcolor{gray!20}{MOODv2 (ours)} & \cellcolor{gray!20}{\revise{\textbf{99.98}}} & \cellcolor{gray!20}{\revise{\textbf{0.06}}} & \cellcolor{gray!20}{\revise{\textbf{100.00}}} & \cellcolor{gray!20}{\revise{\textbf{0.00}}} & \cellcolor{gray!20}{\revise{\textbf{99.94}}} & \cellcolor{gray!20}{\revise{\textbf{0.20}}} & \cellcolor{gray!20}{\revise{\textbf{99.99}}} & \cellcolor{gray!20}{\revise{\textbf{0.01}}} & \cellcolor{gray!20}{\revise{\textbf{99.98}}} & \cellcolor{gray!20}{\revise{\textbf{0.07}}} \\
\midrule
\multirow{10}{*}{ImageNet}
& MSP\cite{baseline_ood} & \revise{71.31} & \revise{77.07} & \revise{90.70} & \revise{43.72} & \revise{60.77} & \revise{90.60} & \revise{84.29} & \revise{61.79} & \revise{76.77} & \revise{68.30} \\
& Energy\cite{energy} & \revise{54.11} & \revise{86.28} & \revise{76.61} & \revise{72.70} & \revise{61.63} & \revise{81.00} & \revise{71.06} & \revise{73.99} & \revise{65.85} & \revise{78.49} \\
& MaxLogit\cite{maxlogit} & \revise{67.22} & \revise{77.98} & \revise{89.88} & \revise{45.57} & \revise{61.68} & \revise{88.60} & \revise{82.73} & \revise{62.52} & \revise{75.37} & \revise{68.67} \\
& KL-Matching\cite{maxlogit} & \revise{82.59} & \revise{67.27} & \revise{87.63} & \revise{69.71} & \revise{66.55} & \revise{88.15} & \revise{84.34} & \revise{74.23} & \revise{80.28} & \revise{74.84} \\
& Residual\cite{vim} & \revise{82.39} & \revise{64.61} & \revise{73.72} & \revise{86.00} & \revise{68.44} & \revise{87.45} & \revise{74.88} & \revise{77.98} & \revise{74.86} & \revise{79.01} \\
& React\cite{react} & \revise{62.09} & \revise{80.47} & \revise{91.20} & \revise{38.74} & \revise{63.66} & \revise{81.00} & \revise{80.43} & \revise{60.41} & \revise{74.34} & \revise{65.15} \\
& Mahalanobis\cite{mahalanobis} & \revise{84.93} & \revise{66.05} & \revise{84.90} & \revise{81.60} & \revise{71.53} & \revise{88.85} & \revise{84.16} & \revise{74.72} & \revise{81.38} & \revise{77.80} \\
& ViM\cite{vim} & \revise{83.51} & \revise{62.71} & \revise{77.75} & \revise{81.72} & \revise{71.04} & \revise{86.60} & \revise{78.31} & \revise{74.55} & \revise{77.65} & \revise{76.40} \\
& \cellcolor{gray!20}{MOODv1\cite{MOOD}} & \cellcolor{gray!20}{\revise{93.01}} & \cellcolor{gray!20}{\revise{30.91}} & \cellcolor{gray!20}{\revise{98.78}} & \cellcolor{gray!20}{\revise{5.89}} & \cellcolor{gray!20}{\revise{86.78}} & \cellcolor{gray!20}{\revise{63.15}} & \cellcolor{gray!20}{\revise{95.46}} & \cellcolor{gray!20}{\revise{26.46}} & \cellcolor{gray!20}{\revise{93.51}} & \cellcolor{gray!20}{\revise{31.60}} \\
& \cellcolor{gray!20}{MOODv2 (ours)} & \cellcolor{gray!20}{\revise{\textbf{94.25}}} & \cellcolor{gray!20}{\revise{\textbf{24.69}}} & \cellcolor{gray!20}{\revise{\textbf{99.59}}} & \cellcolor{gray!20}{\revise{\textbf{1.83}}} & \cellcolor{gray!20}{\revise{\textbf{91.47}}} & \cellcolor{gray!20}{\revise{\textbf{40.80}}} & \cellcolor{gray!20}{\revise{\textbf{97.41}}} & \cellcolor{gray!20}{\revise{\textbf{13.55}}} & \cellcolor{gray!20}{\revise{\textbf{95.68}}} & \cellcolor{gray!20}{\revise{\textbf{20.22}}} \\
\bottomrule
\end{tabular}
\caption{
\revise{Performance of OOD detection methods on ViT-B/16 model with $224\times224$-pixel inputs. All methods are pre-trained on ImageNet-21k and finetuned on ImageNet-1k. ID datasets include CIFAR-10 \cite{cifar} and ImageNet-1k \cite{imagenet}. Both metrics AUROC and FPR95 are in percentage. The best method is emphasized in bold and a gray background indicates our methods.}
}
\label{tab:multi-class}
\end{table*}

\begin{table*}[t]
\small
\centering

\subfloat[AUROC]{
\setlength{\tabcolsep}{2.8mm}
\begin{tabular}{c|cccccccccc|c}
\toprule
\multirow{2}{*}{Methods} & \multicolumn{10}{c|}{ID class} & \multirow{2}{*}{Average} \\
% & 0 & 1 & 2 & 3 & 4 & 5 & 6 & 7 & 8 & 9 & \\
& Plane & Car & Bird & Cat & Deer & Dog & Frog & Horse & Ship & Truck & \\
\midrule
KL-Matching\cite{maxlogit} & \revise{95.35} & \revise{92.04} & \revise{95.18} & \revise{91.26} & \revise{88.11} & \revise{94.66} & \revise{94.99} & \revise{86.52} & \revise{93.61} & \revise{89.37} & \revise{92.11} \\
Residual\cite{vim} & \revise{97.62} & \revise{95.88} & \revise{97.06} & \revise{96.30} & \revise{89.18} & \revise{94.33} & \revise{96.73} & \revise{91.46} & \revise{94.89} & \revise{92.36} & \revise{94.58} \\
Mahalanobis\cite{mahalanobis} & \revise{97.52} & \revise{96.07} & \revise{96.77} & \revise{96.41} & \revise{89.60} & \revise{94.79} & \revise{96.41} & \revise{91.48} & \revise{94.80} & \revise{92.58} & \revise{94.64} \\
ViM\cite{vim} & \revise{97.61} & \revise{96.36} & \revise{97.19} & \revise{96.50} & \revise{88.78} & \revise{94.21} & \revise{96.70} & \revise{91.60} & \revise{94.97} & \revise{92.35} & \revise{94.63} \\
\rowcolor{gray!20}MOODv1\cite{MOOD} & \revise{98.63} & \revise{\textbf{99.33}} & \revise{94.31} & \revise{93.22} & \revise{\textbf{98.11}} & \revise{96.50} & \revise{\textbf{99.25}} & \revise{\textbf{98.96}} & \revise{\textbf{98.76}} & \revise{\textbf{97.82}} & \revise{97.83} \\
\rowcolor{gray!20}MOODv2 (ours) & \revise{\textbf{99.14}} & \revise{99.03} & \revise{\textbf{99.51}} & \revise{\textbf{98.37}} & \revise{97.12} & \revise{\textbf{97.20}} & \revise{98.53} & \revise{98.07} & \revise{98.35} & \revise{96.68} & \revise{\textbf{98.20}} \\
\bottomrule
\end{tabular}
}

\subfloat[FPR95]{
\setlength{\tabcolsep}{2.8mm}
\begin{tabular}{c|cccccccccc|c}
\toprule
\multirow{2}{*}{Methods} & \multicolumn{10}{c|}{ID class} & \multirow{2}{*}{Average} \\
% & 0 & 1 & 2 & 3 & 4 & 5 & 6 & 7 & 8 & 9 & \\
& Plane & Car & Bird & Cat & Deer & Dog & Frog & Horse & Ship & Truck & \\
\midrule
KL-Matching\cite{maxlogit} & \revise{23.60} & \revise{32.60} & \revise{22.32} & \revise{42.92} & \revise{46.26} & \revise{24.30} & \revise{24.97} & \revise{46.74} & \revise{25.32} & \revise{40.53} & \revise{32.96} \\
Residual\cite{vim} & \revise{12.06} & \revise{25.58} & \revise{16.71} & \revise{21.17} & \revise{48.33} & \revise{22.12} & \revise{17.42} & \revise{36.72} & \revise{17.30} & \revise{30.76} & \revise{24.82} \\
Mahalanobis\cite{mahalanobis} & \revise{12.59} & \revise{25.72} & \revise{18.92} & \revise{21.48} & \revise{48.44} & \revise{20.59} & \revise{19.20} & \revise{38.02} & \revise{17.47} & \revise{30.93} & \revise{25.34} \\
ViM\cite{vim} & \revise{12.43} & \revise{24.83} & \revise{15.77} & \revise{20.13} & \revise{48.68} & \revise{21.77} & \revise{17.63} & \revise{36.63} & \revise{17.60} & \revise{30.78} & \revise{24.63} \\
\rowcolor{gray!20}MOODv1\cite{MOOD} & \revise{7.59} & \revise{5.04} & \revise{2.47} & \revise{\textbf{7.49}} & \revise{15.63} & \revise{\textbf{10.96}} & \revise{11.37} & \revise{13.09} & \revise{10.06} & \revise{19.62} & \revise{10.33} \\
\rowcolor{gray!20}MOODv2 (ours) & \revise{\textbf{4.82}} & \revise{\textbf{4.50}} & \revise{\textbf{1.79}} & \revise{8.80} & \revise{\textbf{15.59}} & \revise{11.00} & \revise{\textbf{8.46}} & \revise{\textbf{12.43}} & \revise{\textbf{8.60}} & \revise{\textbf{18.96}} & \revise{\textbf{9.49}} \\
\bottomrule
\end{tabular}
}

\caption{
\revise{Performance of OOD detection methods on ViT-B/16 model with $224\times224$-pixel inputs. All methods are pre-trained on ImageNet-21k and finetuned on ImageNet-1k. We perform each category of CIFAR-10 \cite{cifar} as the ID dataset and other classes as OOD datasets. We report the average results across OOD classes of each ID class. Both metrics AUROC and FPR95 are in percentage. The best method is emphasized in bold and a gray background indicates our methods.}
} % The detailed class-wize performance is in the Appendix.
\label{tab:one-class}
\end{table*}

\section{Experiments}\label{sec:exp}

\label{sec:exp}
\revise{In this section, we conduct a thorough comparison of our algorithm with the latest OOD detection methods. We employ the ViT-B/16 model, pre-trained on ImageNet-21K and fine-tuned on ImageNet-1K at a resolution of $224\times224$. }

\vspace{2mm}\noindent\revise{\textbf{ID/OOD Datasets.} We select CIFAR-10 \cite{cifar} and ImageNet-1K~\cite{imagenet} as the ID datasets. Following established procedures~\cite{vim}, for estimating the principal space of ImageNet, we randomly sample $200,000$ images from the training set. Our experiments include the following OOD datasets:
\begin{enumerate}
    \item OpenImage-O is a newly collected large-scale OOD dataset~\cite{openimages_o}.
    \item Texture~\cite{cimpoi14describing} comprises natural textural images, with four overlapping categories (\emph{bubbly, honeycombed, cobwebbed, spiraled}) removed since they coincide with ImageNet.
    \item iNaturalist~\cite{van2018inaturalist} is a fine-grained species classification dataset, and we use a specific subset from previous works ~\cite{huang2021mos}.
    \item ImageNet-O~\cite{hendrycks2021natural} contains images that are adversarially filtered to challenge OOD detectors. 
\end{enumerate} 
}

\vspace{2mm}\noindent\revise{\textbf{Evaluation Metrics.} 
We report two commonly used evaluation metrics AUROC and FPR95. The AUROC is a threshold-free metric, indicating the area under the receiver operating characteristic curve, with a higher value denoting better detection performance. FPR95, or FPR at TPR95, stands for the false positive rate when the true positive rate is 95\%, and a smaller FPR95 is preferable. Both metrics are expressed as percentages.}

\vspace{2mm}\noindent\revise{\textbf{Baseline Methods.} 
Following previous works \cite{vim}, we compare MOODv2 with the baseline algorithms that do not require fine-tuning including {MSP} \cite{baseline_ood}, {Energy} \cite{energy}, {ODIN} \cite{odin}, {MaxLogit} \cite{maxlogit}, {KL Matching} \cite{maxlogit}, {Residual}, {ReAct} \cite{react}, and {Mahalanobis} \cite{mahalanobis}. }

%--------------------------------------visualization----------------
\begin{figure*}[tp]
  \centering
    \includegraphics[width=\textwidth]{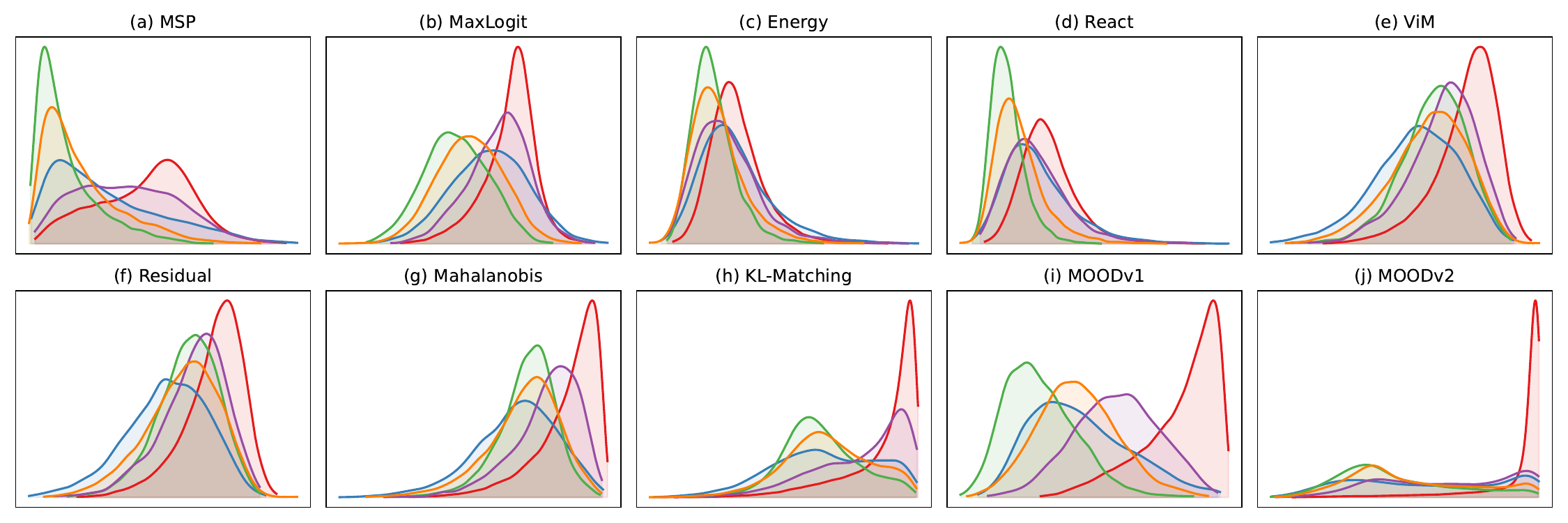}
    \caption{\revise{The distribution curves of OOD score functions for ID and OOD datasets obtained using various mainstream methods, including {MSP} \cite{baseline_ood}, {Energy} \cite{energy}, {ODIN} \cite{odin}, {MaxLogit} \cite{maxlogit}, {KL Matching} \cite{maxlogit}, {Residual} \cite{vim}, {ReAct} \cite{react}, {Mahalanobis} \cite{mahalanobis} and {ViM} \cite{vim}. The red line indicates the ID dataset ImageNet \cite{imagenet}; the blue line indicates Texture \cite{dtd}; the green line indicates iNaturalist \cite{inaturalist}; the purple line indicates ImageNet-O \cite{imagenet_o}; the orange line indicates OpenImage-O \cite{openimages_o}}}
  \label{fig:distribution}
\end{figure*}

\subsection{One-Class OOD Detection}
\label{sec:1class}
We start with the one-class OOD detection. For a given multi-class dataset of $N_c$ classes, we conduct $N_c$ one-class OOD tasks, where each task regards one of the classes as in-distribution and the remaining classes as out-of-distribution. We run our experiments on CIFAR-10 \cite{cifar}. \Cref{tab:one-class} summarizes the average results across OOD classes of each ID class and the detailed class-wize performance is in the appendix.

\revise{It's worth noting that all methods were pre-trained on ImageNet-21k and fine-tuned on ImageNet-1k, which may have had some influence on the results to varying degrees. Nevertheless, we ensure consistent training strategies for all methods to ensure a fair comparison. Experimental results have demonstrated that MOODv2 achieves significant improvements across all ID classes even without fine-tuning the ID dataset. Notably, we achieved a remarkable 3.56\% increase in the AUROC, reaching 98.20\%, while simultaneously reducing the FPR95 by 15.14\% to achieve an impressive 9.49\%.}

\subsection{Multi-Class OOD Detection} 
\label{sec:multi-class}
\revise{For multi-class OOD Detection, we assume that ID samples are from a multi-class dataset, either CIFAR-10 \cite{cifar} or ImageNet \cite{imagenet}. They are tested on external datasets as out-of-distribution, including OpenImage-O \cite{openimages_o}, Texture \cite{cimpoi14describing}, iNaturalist \cite{van2018inaturalist} and ImageNet-O \cite{hendrycks2021natural}.}

\revise{Results are shown in \cref{tab:multi-class}. MOODv2 delivers outstanding results on CIFAR-10, achieving an impressive AUROC of 99.98\% (0.35\% enhancement) and the FPR95 reaches an astonishingly low rate of 0.07\%, marking a substantial 95\% reduction compared to the prior SOTA (1.39\%). On ImageNet, MOODv2 also exhibited significant improvements, showcasing a remarkable 14.30\% increase in AUROC, resulting in 95.68\%. Additionally, the FPR95 saw a substantial reduction of 44.93\%, reaching 20.22\%.}

\revise{In \cref{fig:distribution}, we illustrate the distribution curves of OOD scores for ID and OOD datasets using various mainstream methods. A smaller overlap between ID and OOD data indicates superior OOD detection performance, while a larger overlap signifies weaker detection results. The ID curve (in red) for MOODv2 features a distinct peak at a higher position, resulting in minimal overlap with other OOD data, indicating a notable OOD detection capability. This success can be attributed to the high-quality ID feature representation.}

\section{Conclusion}\label{sec:conclusion}
\revise{In our work, we focus on the critical aspect of effective out-of-distribution (OOD) detection, which involves acquiring a robust in-distribution (ID) representation that distinguishes it from OOD samples. We conduct comprehensive experiments with distinct pretraining tasks and employ various OOD score functions. The findings indicate that feature representations pre-trained through reconstruction significantly enhance performance and reduce the performance gap among different score functions. This implies that even simple score functions can perform as well as complex ones when utilizing reconstruction-based pretext tasks. These findings hold promise for further development in OOD detection. Ultimately, we introduce the MOODv2 OOD detection framework, employing the masked image modeling pretext task, which achieves a remarkable 14.30\% increase in AUROC, reaching 95.68\% on ImageNet, and substantially improving CIFAR-10 to 99.98\%.}

\ifCLASSOPTIONcaptionsoff
  \newpage
\fi

\bibliographystyle{IEEEtran}
\bibliography{egbib}

\begin{IEEEbiography}
 [{\includegraphics[height=1.10in,clip,keepaspectratio]{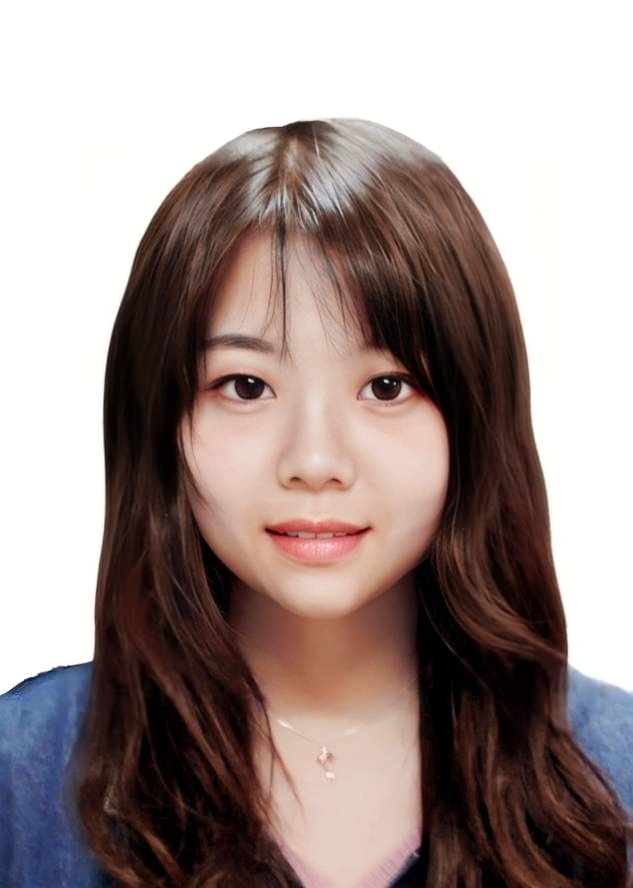}}]
 {Jingyao Li} received the B.Eng. degree from Xi'an Jiaotong University. She is currently a Ph.D. student at Department of Computer Science and Engineering of the Chinese University of Hong Kong (CUHK), under the supervision of Prof. Jiaya Jia. She serves as a reviewer for CVPR, ECCV, ICCV and etc. Her research interests include computer vision and large language models.
\end{IEEEbiography}

\begin{IEEEbiography}
 [{\includegraphics[height=1.25in,clip,keepaspectratio]{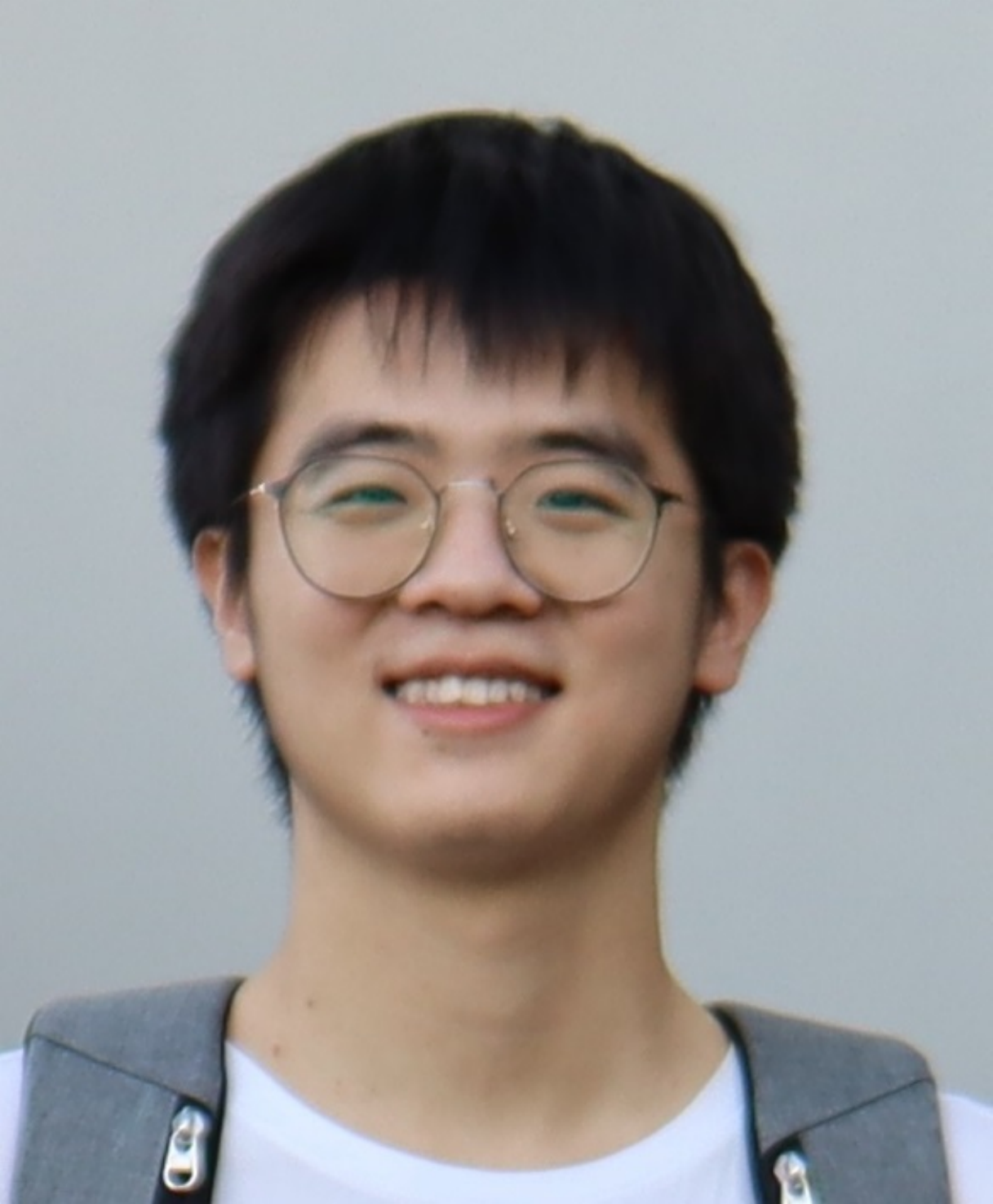}}] {Pengguang Chen} received the B.Eng. degree in Computer Science from Nanjing University and the Ph.D. degree from the Chinese University of Hong Kong (CUHK), under the supervision of Prof. Jiaya Jia. He is currently a researcher in SmartMore. He serves as a reviewer for CVPR, ICCV, ECCV, TPAMI. His research interests include neural architecture search, self-supervised learning, knowledge distillation and semantic segmentation.
\end{IEEEbiography}

\begin{IEEEbiography}[{\includegraphics[width=1in,height=1.10in,clip,keepaspectratio]{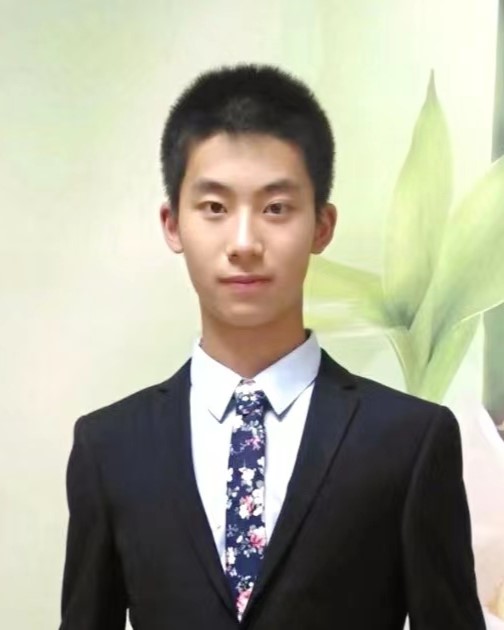}}]{Shaozuo Yu} is a Ph.D. student at Department of Computer Science and Engineering of the Chinese University of Hong Kong. He served as a program chair of the workshop and challenge on “Out-of-Distribution Generalization in Computer Vision” at ECCV’22. He served as a reviewer for CVPR, Neurips, and ICML. His research interests include multimodality, generative models, and robust vision.
\end{IEEEbiography}

\begin{IEEEbiography}
[{\includegraphics[width=1in,height=1.25in,clip,keepaspectratio]{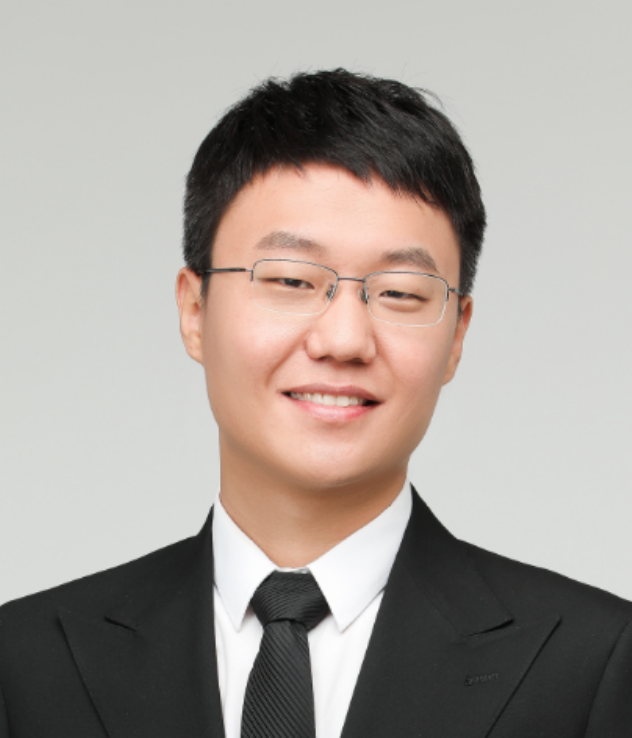}}]{Shu Liu} now serves as Co-Founder and Technical Head in SmartMore. He received the BS degree from Huazhong University of Science and Technology and the PhD degree from the Chinese University of Hong Kong. He was the winner of 2017 COCO Instance Segmentation Competition and received the Outstanding Reviewer of ICCV in 2019. He continuously served as a reviewer for TPAMI, CVPR, ICCV, NIPS, ICLR and etc. His research interests lie in deep learning and computer vision. 
  He is a member of IEEE.
 \end{IEEEbiography} 

\begin{IEEEbiography}
[{\includegraphics[width=1in,height=1.25in,clip,keepaspectratio]{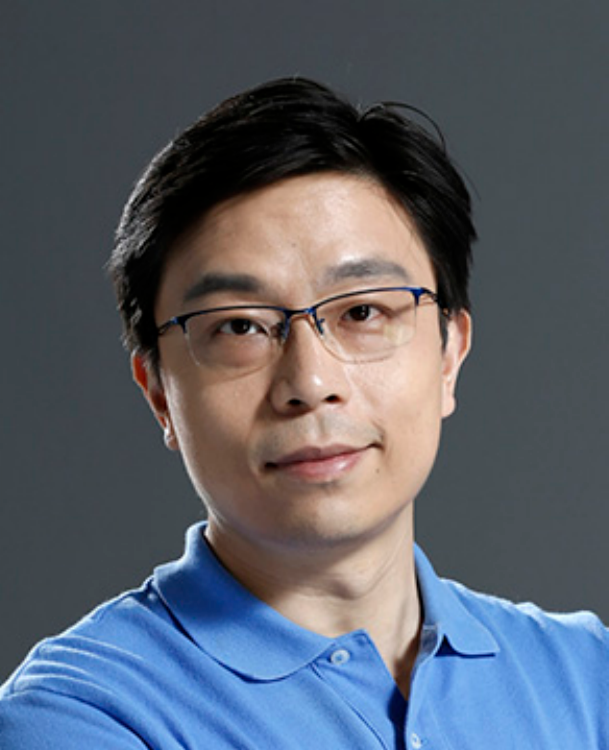}}]{Jiaya Jia} received the Ph.D.~degree in Computer Science from Hong Kong University of Science and Technology in 2004 and is currently a full professor in Department of Computer Science and Engineering at the Chinese University of Hong Kong (CUHK). He assumes the position of Associate Editor-in-Chief of IEEE Transactions on Pattern Analysis and Machine Intelligence (TPAMI) and is in the editorial board of International Journal of Computer Vision (IJCV). He continuously served as area chairs for ICCV, CVPR, AAAI, ECCV, and several other conferences for the organization. He was on program committees of major conferences in graphics and computational imaging, including ICCP, SIGGRAPH, and SIGGRAPH Asia. He is a Fellow of the IEEE. 
\end{IEEEbiography}

\appendices

This supplementary material includes visualization of distribution curves, multi-class and one-class OOD detection results on CIFAR-10, etc., which are not included in the main paper due to page limitations. 

\section{Distribution curves}
For more comprehensive insights, we offer visual representations of distribution curves for OOD scores on both ID and OOD datasets in \cref{fig:distribution}. A narrower overlap between ID and OOD data signifies superior OOD detection performance, whereas a wider overlap indicates weaker detection results. The ID curve, depicted in red, for the fine-tuned BEiT series \cite{beit} models, exhibits a distinctive peak at a higher position. This leads to minimal overlap with other OOD data, highlighting a remarkable OOD detection capability. This accomplishment can be attributed to the high-quality ID feature representation derived from masked image modeling.

\section{Details of Results on CIFAR-10}
\subsection{Multi-class OOD Detection}

We employ CIFAR-10 \cite{cifar} as the in-distribution dataset and evaluate pre-task texts on multiple challenging unnatural out-of-distribution datasets, including OpenImage-O \cite{openimages_o}, Texture \cite{dtd}, iNaturalist \cite{inaturalist}, and ImageNet-O \cite{imagenet_o}. Extensive validations with various pretraining methods and OOD score functions including {MSP} \cite{baseline_ood}, {Energy} \cite{energy}, {ODIN} \cite{odin}, {MaxLogit} \cite{maxlogit}, {KL Matching} \cite{maxlogit}, {Residual} \cite{vim}, {ReAct} \cite{react}, {Mahalanobis} \cite{mahalanobis} and {ViM} \cite{vim}. Results are in \cref{tab:multi-class-cifar10-ablation}. Our approach attains an impressive AUROC of 99.99\% while concurrently reducing the FPR95 to a mere 0.03\%.

\subsection{One-class OOD Detection}
We perform one-class OOD detection. In the context of a multi-class dataset with $N_c$ classes, we conduct $N_c$ one-class OOD tasks. Each task treats one of the classes as in-distribution and the remaining classes as out-of-distribution. Our experiments are conducted on CIFAR-10 \cite{cifar} and provide the detailed class-wise performance of mainstream methods including KL-Marching (\cref{tab:one-class-detailed-kl}), Residual (\cref{tab:one-class-detailed-residual}), Mahalanobis (\cref{tab:one-class-detailed-distance}) and ViM (\cref{tab:one-class-detailed-vim}).

It's important to note that all methods were pre-trained on ImageNet-21k and subsequently fine-tuned on ImageNet-1k, which might have influenced the results to varying degrees. However, we ensure consistent training strategies for all methods to maintain a fair comparison. The experimental results demonstrate that MOODv2 achieves significant improvements across all ID classes, even without fine-tuning the ID dataset. Notably, we achieved a remarkable 3.56\% increase in the state-of-the-art AUROC, reaching 98.20\%, while simultaneously reducing FPR95 by 15.14\%, achieving an impressive 9.49\%.

\begin{figure*}[tp]
  \centering
    \includegraphics[width=\textwidth]{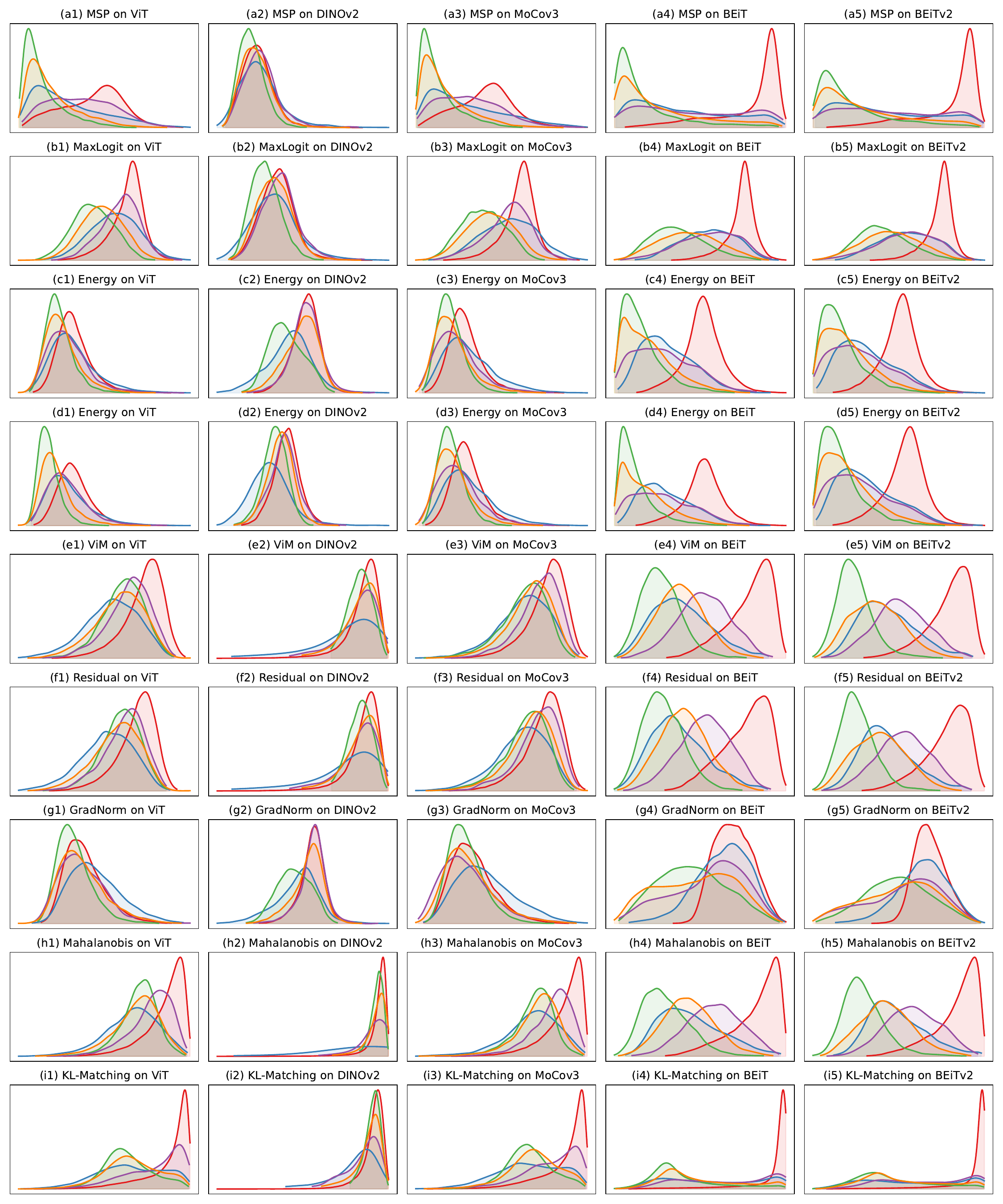}
    \caption{The distribution curves of OOD scores for ID and OOD datasets obtained using various mainstream methods, including {MSP} \cite{baseline_ood}, {Energy} \cite{energy}, {ODIN} \cite{odin}, {MaxLogit} \cite{maxlogit}, {KL Matching} \cite{maxlogit}, {Residual} \cite{vim}, {ReAct} \cite{react}, {Mahalanobis} \cite{mahalanobis} and {ViM} \cite{vim}. The red line indicates the ID dataset ImageNet \cite{imagenet}; the blue line indicates Texture \cite{dtd}; the green line indicates iNaturalist \cite{inaturalist}; the purple line indicates ImageNet-O \cite{imagenet_o}; the orange line indicates OpenImage-O \cite{openimages_o}. Pretrained models include classification task \cite{vit}, MoCov3 \cite{mocov3}, DINOv2 \cite{dinov2}, BEiTv2 \cite{beitv2} and BEiT \cite{beit}.}
  \label{fig:distribution}
\end{figure*}

% \input{tables/mim-imagenet-detailed}

%--------------------------------------multi-class-----------------------------------
\begin{table*}[t!]
\small
\centering
\setlength{\tabcolsep}{0.6mm}
\begin{tabular}{c|c|ccccccccccc}
\toprule
\multirow{2}{*}{Methods} & \multirow{2}{*}{Models} & \multicolumn{2}{c}{Texture} & \multicolumn{2}{c}{iNaturalist} & \multicolumn{2}{c}{ImageNet-O} & \multicolumn{2}{c}{OpenImage-O} & \multicolumn{2}{c}{Average} \\
& & AUROC$\uparrow$ & FPR95$\downarrow$ & AUROC$\uparrow$ & FPR95$\downarrow$ & AUROC$\uparrow$ & FPR95$\downarrow$ & AUROC$\uparrow$ & FPR95$\downarrow$ & AUROC$\uparrow$ & FPR95$\downarrow$ \\
\midrule
\multirow{5}{*}{MSP\cite{baseline_ood}}
& ViT\cite{vit} & 45.67 & 95.17 & 71.07 & 81.76 & 32.52 & 98.85 & 59.74 & 91.45 & 52.25 & 91.81 \\
& MoCov3\cite{mocov3} & 37.11 & 97.64 & 64.60 & 89.69 & 31.52 & 98.45 & 55.90 & 93.53 & 47.28 & 94.83 \\
& DINOv2\cite{dinov2} & \textbf{70.37} & \textbf{58.33} & \textbf{87.10} & \textbf{37.77} & \textbf{70.61} & \textbf{63.25} & \textbf{78.27} & \textbf{51.58} & \textbf{76.59} & \textbf{52.73} \\
& \cellcolor{gray!20}{BEiTv2\cite{beitv2}} & \cellcolor{gray!20}{57.67} & \cellcolor{gray!20}{88.31} & \cellcolor{gray!20}{82.53} & \cellcolor{gray!20}{55.54} & \cellcolor{gray!20}{52.06} & \cellcolor{gray!20}{89.55} & \cellcolor{gray!20}{72.72} & \cellcolor{gray!20}{70.71} & \cellcolor{gray!20}{66.24} & \cellcolor{gray!20}{76.03} \\
& \cellcolor{gray!20}{BEiT\cite{beit} } & \cellcolor{gray!20}{51.64} & \cellcolor{gray!20}{91.09} & \cellcolor{gray!20}{73.85} & \cellcolor{gray!20}{74.02} & \cellcolor{gray!20}{47.82} & \cellcolor{gray!20}{90.75} & \cellcolor{gray!20}{65.40} & \cellcolor{gray!20}{80.81} & \cellcolor{gray!20}{59.68} & \cellcolor{gray!20}{84.17} \\
\midrule
\multirow{5}{*}{Energy\cite{energy}}
& ViT\cite{vit} & 31.16 & 97.89 & 48.95 & 97.92 & 37.22 & 97.85 & 45.29 & 96.36 & 40.65 & 97.50 \\
& MoCov3\cite{mocov3} & 24.97 & 98.93 & 44.74 & 98.29 & 38.06 & 95.45 & 43.90 & 95.49 & 37.92 & 97.04 \\
& DINOv2\cite{dinov2} & \textbf{86.73} & \textbf{28.16} & \textbf{91.43} & \textbf{20.27} & \textbf{68.97} & \textbf{62.75} & 73.66 & 53.40 & \textbf{80.20} & \textbf{41.15} \\
& \cellcolor{gray!20}{BEiTv2\cite{beitv2}} & \cellcolor{gray!20}{63.35} & \cellcolor{gray!20}{82.64} & \cellcolor{gray!20}{88.52} & \cellcolor{gray!20}{38.21} & \cellcolor{gray!20}{66.24} & \cellcolor{gray!20}{72.30} & \cellcolor{gray!20}{\textbf{81.69}} & \cellcolor{gray!20}{\textbf{51.80}} & \cellcolor{gray!20}{74.95} & \cellcolor{gray!20}{61.24} \\
& \cellcolor{gray!20}{BEiT\cite{beit} } & \cellcolor{gray!20}{52.98} & \cellcolor{gray!20}{88.53} & \cellcolor{gray!20}{81.84} & \cellcolor{gray!20}{54.55} & \cellcolor{gray!20}{59.99} & \cellcolor{gray!20}{77.20} & \cellcolor{gray!20}{73.61} & \cellcolor{gray!20}{65.08} & \cellcolor{gray!20}{67.10} & \cellcolor{gray!20}{71.34} \\
\midrule
\multirow{5}{*}{MaxLogit\cite{maxlogit}}
& ViT\cite{vit} & 41.21 & 95.95 & 67.83 & 86.04 & 32.58 & 98.80 & 56.64 & 92.94 & 49.56 & 93.43 \\
& MoCov3\cite{mocov3} & 32.94 & 98.22 & 61.79 & 92.21 & 31.65 & 98.45 & 53.32 & 94.26 & 44.92 & 95.78 \\
& DINOv2\cite{dinov2} & \textbf{76.80} & \textbf{45.06} & \textbf{91.96} & \textbf{22.66} & \textbf{72.49} & \textbf{57.95} & \textbf{80.36} & \textbf{44.75} & \textbf{80.40} & \textbf{42.61} \\
& \cellcolor{gray!20}{BEiTv2\cite{beitv2}} & \cellcolor{gray!20}{60.51} & \cellcolor{gray!20}{85.85} & \cellcolor{gray!20}{86.05} & \cellcolor{gray!20}{47.39} & \cellcolor{gray!20}{59.14} & \cellcolor{gray!20}{83.90} & \cellcolor{gray!20}{77.47} & \cellcolor{gray!20}{62.79} & \cellcolor{gray!20}{70.79} & \cellcolor{gray!20}{69.98} \\
& \cellcolor{gray!20}{BEiT\cite{beit} } & \cellcolor{gray!20}{51.94} & \cellcolor{gray!20}{89.90} & \cellcolor{gray!20}{77.92} & \cellcolor{gray!20}{66.72} & \cellcolor{gray!20}{52.95} & \cellcolor{gray!20}{87.10} & \cellcolor{gray!20}{69.16} & \cellcolor{gray!20}{75.24} & \cellcolor{gray!20}{62.99} & \cellcolor{gray!20}{79.74} \\
\midrule
\multirow{5}{*}{KL-Matching\cite{maxlogit}}
& ViT\cite{vit} & 98.00 & 10.64 & 94.23 & 35.86 & 92.99 & 32.40 & 94.68 & 27.92 & 94.97 & 26.71 \\
& MoCov3\cite{mocov3} & 97.61 & 13.97 & 94.65 & 35.51 & 92.05 & 38.25 & 94.22 & 33.48 & 94.64 & 30.30 \\
& DINOv2\cite{dinov2} & \textbf{98.05} & \textbf{8.74} & \textbf{98.95} & \textbf{5.32} & \textbf{97.29} & \textbf{12.35} & \textbf{96.99} & \textbf{13.55} & \textbf{97.82} & \textbf{9.99} \\
& \cellcolor{gray!20}{BEiTv2\cite{beitv2}} & \cellcolor{gray!20}{97.41} & \cellcolor{gray!20}{14.98} & \cellcolor{gray!20}{91.78} & \cellcolor{gray!20}{50.90} & \cellcolor{gray!20}{93.21} & \cellcolor{gray!20}{35.90} & \cellcolor{gray!20}{92.28} & \cellcolor{gray!20}{43.17} & \cellcolor{gray!20}{93.67} & \cellcolor{gray!20}{36.24} \\
& \cellcolor{gray!20}{BEiT\cite{beit} } & \cellcolor{gray!20}{97.83} & \cellcolor{gray!20}{12.71} & \cellcolor{gray!20}{95.14} & \cellcolor{gray!20}{32.28} & \cellcolor{gray!20}{93.52} & \cellcolor{gray!20}{35.00} & \cellcolor{gray!20}{94.84} & \cellcolor{gray!20}{31.13} & \cellcolor{gray!20}{95.33} & \cellcolor{gray!20}{27.78} \\
\midrule
\multirow{5}{*}{Residual\cite{vim}}
& ViT\cite{vit} & 99.91 & 0.21 & 99.68 & 0.45 & 99.36 & 2.85 & 99.42 & 2.46 & 99.59 & 1.49 \\
& MoCov3\cite{mocov3} & 99.90 & 0.25 & 99.87 & 0.09 & 99.22 & 3.85 & 99.59 & 1.31 & 99.65 & 1.38 \\
& DINOv2\cite{dinov2} & 99.98 & 0.04 & 100.00 & 0.01 & \textbf{99.99} & \textbf{0.05} & 99.97 & 0.18 & 99.98 & 0.07 \\
& \cellcolor{gray!20}{BEiTv2\cite{beitv2}} & \cellcolor{gray!20}{99.98} & \cellcolor{gray!20}{0.04} & \cellcolor{gray!20}{100.00} & \cellcolor{gray!20}{\textbf{0.00}} & \cellcolor{gray!20}{99.79} & \cellcolor{gray!20}{0.90} & \cellcolor{gray!20}{99.92} & \cellcolor{gray!20}{0.27} & \cellcolor{gray!20}{99.92} & \cellcolor{gray!20}{0.30} \\
& \cellcolor{gray!20}{BEiT\cite{beit} } & \cellcolor{gray!20}{\textbf{99.99}} & \cellcolor{gray!20}{\textbf{0.02}} & \cellcolor{gray!20}{\textbf{100.00}} & \cellcolor{gray!20}{\textbf{0.00}} & \cellcolor{gray!20}{99.96} & \cellcolor{gray!20}{0.10} & \cellcolor{gray!20}{\textbf{99.99}} & \cellcolor{gray!20}{\textbf{0.01}} & \cellcolor{gray!20}{\textbf{99.99}} & \cellcolor{gray!20}{\textbf{0.03}} \\
\midrule
\multirow{5}{*}{React\cite{react}}
& ViT\cite{vit} & 35.97 & 96.26 & 69.01 & 87.91 & 36.65 & 97.75 & 54.14 & 93.11 & 48.94 & 93.76 \\
& MoCov3\cite{mocov3} & 25.74 & 98.90 & 46.11 & 98.29 & 37.60 & 95.55 & 44.63 & 95.46 & 38.52 & 97.05 \\
& DINOv2\cite{dinov2} & \textbf{68.00} & \textbf{61.94} & 60.58 & 82.56 & 40.71 & 90.10 & 47.60 & 88.88 & 54.22 & 80.87 \\
& \cellcolor{gray!20}{BEiTv2\cite{beitv2}} & \cellcolor{gray!20}{62.81} & \cellcolor{gray!20}{82.71} & \cellcolor{gray!20}{\textbf{91.59}} & \cellcolor{gray!20}{\textbf{28.05}} & \cellcolor{gray!20}{\textbf{65.37}} & \cellcolor{gray!20}{\textbf{72.95}} & \cellcolor{gray!20}{\textbf{82.43}} & \cellcolor{gray!20}{\textbf{49.49}} & \cellcolor{gray!20}{\textbf{75.55}} & \cellcolor{gray!20}{\textbf{58.30}} \\
& \cellcolor{gray!20}{BEiT\cite{beit} } & \cellcolor{gray!20}{53.27} & \cellcolor{gray!20}{87.91} & \cellcolor{gray!20}{82.09} & \cellcolor{gray!20}{53.87} & \cellcolor{gray!20}{59.64} & \cellcolor{gray!20}{77.50} & \cellcolor{gray!20}{73.73} & \cellcolor{gray!20}{64.84} & \cellcolor{gray!20}{67.18} & \cellcolor{gray!20}{71.03} \\
\midrule
\multirow{5}{*}{Mahalanobis\cite{mahalanobis}}
& ViT\cite{vit} & 99.77 & 0.60 & 99.39 & 1.11 & 98.93 & 4.90 & 99.14 & 3.26 & 99.31 & 2.47 \\
& MoCov3\cite{mocov3} & 99.78 & 0.78 & 99.71 & 0.45 & 98.61 & 7.65 & 99.31 & 2.48 & 99.35 & 2.84 \\
& DINOv2\cite{dinov2} & 99.98 & 0.06 & 100.00 & \textbf{0.00} & \textbf{99.99} & \textbf{0.00} & 99.97 & 0.16 & \textbf{99.99} & 0.05 \\
& \cellcolor{gray!20}{BEiTv2\cite{beitv2}} & \cellcolor{gray!20}{99.95} & \cellcolor{gray!20}{0.06} & \cellcolor{gray!20}{99.99} & \cellcolor{gray!20}{0.02} & \cellcolor{gray!20}{99.61} & \cellcolor{gray!20}{1.90} & \cellcolor{gray!20}{99.82} & \cellcolor{gray!20}{0.77} & \cellcolor{gray!20}{99.84} & \cellcolor{gray!20}{0.69} \\
& \cellcolor{gray!20}{BEiT\cite{beit} } & \cellcolor{gray!20}{\textbf{99.99}} & \cellcolor{gray!20}{\textbf{0.00}} & \cellcolor{gray!20}{\textbf{100.00}} & \cellcolor{gray!20}{\textbf{0.00}} & \cellcolor{gray!20}{99.96} & \cellcolor{gray!20}{0.05} & \cellcolor{gray!20}{\textbf{99.98}} & \cellcolor{gray!20}{\textbf{0.05}} & \cellcolor{gray!20}{99.98} & \cellcolor{gray!20}{\textbf{0.03}} \\
\midrule
\multirow{5}{*}{ViM\cite{vim}}
& ViT\cite{vit} & 99.91 & 0.23 & 99.72 & 0.38 & 99.38 & 2.65 & 99.49 & 2.31 & 99.63 & 1.39 \\
& MoCov3\cite{mocov3} & 99.93 & 0.16 & 99.92 & 0.03 & 99.40 & 2.75 & 99.69 & 1.03 & 99.73 & 0.99 \\
& DINOv2\cite{dinov2} & \textbf{99.98} & \textbf{0.04} & 100.00 & 0.01 & \textbf{99.99} & \textbf{0.05} & 99.97 & 0.18 & \textbf{99.98} & 0.07 \\
& \cellcolor{gray!20}{BEiTv2\cite{beitv2}} & \cellcolor{gray!20}{99.95} & \cellcolor{gray!20}{0.14} & \cellcolor{gray!20}{100.00} & \cellcolor{gray!20}{0.01} & \cellcolor{gray!20}{99.61} & \cellcolor{gray!20}{1.60} & \cellcolor{gray!20}{99.93} & \cellcolor{gray!20}{0.28} & \cellcolor{gray!20}{99.87} & \cellcolor{gray!20}{0.51} \\
& \cellcolor{gray!20}{BEiT\cite{beit} } & \cellcolor{gray!20}{99.98} & \cellcolor{gray!20}{0.06} & \cellcolor{gray!20}{\textbf{100.00}} & \cellcolor{gray!20}{\textbf{0.00}} & \cellcolor{gray!20}{99.94} & \cellcolor{gray!20}{0.20} & \cellcolor{gray!20}{\textbf{99.99}} & \cellcolor{gray!20}{\textbf{0.01}} & \cellcolor{gray!20}{99.98} & \cellcolor{gray!20}{\textbf{0.07}} \\
\midrule
\multirow{5}{*}{Best}
& ViT\cite{vit} & 99.91 & 0.21 & 99.72 & 0.38 & 99.38 & 2.65 & 99.49 & 2.31 & 99.63 & 1.39 \\
& MoCov3\cite{mocov3} & 99.93 & 0.16 & 99.92 & 0.03 & 99.40 & 2.75 & 99.69 & 1.03 & 99.73 & 0.99 \\
& DINOv2\cite{dinov2} & 99.98 & 0.04 & 100.00 & \textbf{0.00} & \textbf{99.99} & \textbf{0.00} & 99.97 & 0.16 & 99.99 & 0.05 \\
& \cellcolor{gray!20}{BEiTv2\cite{beitv2}} & \cellcolor{gray!20}{99.98} & \cellcolor{gray!20}{0.04} & \cellcolor{gray!20}{100.00} & \cellcolor{gray!20}{\textbf{0.00}} & \cellcolor{gray!20}{99.79} & \cellcolor{gray!20}{0.90} & \cellcolor{gray!20}{99.93} & \cellcolor{gray!20}{0.27} & \cellcolor{gray!20}{99.92} & \cellcolor{gray!20}{0.30} \\
& \cellcolor{gray!20}{BEiT\cite{beit} } & \cellcolor{gray!20}{\textbf{99.99}} & \cellcolor{gray!20}{\textbf{0.00}} & \cellcolor{gray!20}{\textbf{100.00}} & \cellcolor{gray!20}{\textbf{0.00}} & \cellcolor{gray!20}{99.96} & \cellcolor{gray!20}{0.05} & \cellcolor{gray!20}{\textbf{99.99}} & \cellcolor{gray!20}{\textbf{0.01}} & \cellcolor{gray!20}{\textbf{99.99}} & \cellcolor{gray!20}{\textbf{0.03}} \\
\bottomrule
\end{tabular}
\caption{
AUROC (\%) of OOD detection methods. 
The ID dataset is CIFAR-10 \cite{cifar}, and the OOD datasets are OpenImage-O \cite{openimages_o}, Texture \cite{dtd}, iNaturalist \cite{inaturalist}, and ImageNet-O \cite{imagenet_o}.
The pre-text tasks include classical classification task \cite{vit}, contrastive learning tasks MoCov3 \cite{mocov3} and DINOv2 \cite{dinov2}, and masked image modeling tasks BEiT \cite{beit} and BEiT \cite{beitv2}. All pre-text tasks are performed on ImageNet-21k.
Both metrics AUROC and FPR95 are in percentage.
A pre-trained ViT-B/16 model with $224\times224$-pixel inputs is tested.
The best method is emphasized in bold and a gray background indicates our choice.
}

\label{tab:multi-class-cifar10-ablation}
\end{table*}
% -------------------------------------------------------------
\begin{table*}[t]
\small
\centering
\setlength{\tabcolsep}{2mm}
\begin{tabular}{c|c|cccccccccc|c}
\toprule
\multirow{2}{*}{Models} & \multirow{2}{*}{ID class} & \multicolumn{10}{c|}{OOD class} & \multirow{2}{*}{Average} \\
 &  & 0 & 1 & 2 & 3 & 4 & 5 & 6 & 7 & 8 & 9 & \\
\midrule
\multirow{10}{*}{ViT\cite{vit}} & 0 & - & 81.29 & 90.44 & 90.62 & 86.42 & 95.45 & 93.28 & 91.15 & 67.66 & 66.06 & 84.71 \\
& 1 & 98.54 & - & 99.58 & 99.33 & 99.57 & 99.69 & 99.82 & 96.16 & 96.00 & 85.23 & 97.10 \\
& 2 & 90.41 & 97.50 & - & 87.18 & 76.75 & 95.02 & 90.45 & 81.47 & 96.76 & 98.16 & 90.41 \\
& 3 & 94.14 & 94.49 & 91.32 & - & 77.12 & 78.07 & 88.59 & 82.25 & 98.25 & 85.24 & 87.72 \\
& 4 & 96.15 & 98.55 & 92.46 & 89.63 & - & 95.80 & 94.34 & 66.06 & 98.45 & 96.45 & 91.99 \\
& 5 & 98.38 & 98.19 & 96.45 & 78.99 & 86.87 & - & 97.45 & 68.19 & 99.48 & 95.96 & 91.11 \\
& 6 & 96.68 & 97.44 & 92.05 & 88.20 & 83.14 & 95.27 & - & 96.14 & 95.03 & 97.92 & 93.54 \\
& 7 & 96.32 & 97.72 & 97.36 & 92.15 & 85.78 & 94.45 & 98.03 & - & 94.20 & 98.45 & 94.94 \\
& 8 & 89.88 & 71.86 & 97.40 & 95.86 & 98.82 & 98.45 & 93.19 & 98.04 & - & 80.89 & 91.60 \\
& 9 & 97.62 & 91.34 & 99.60 & 99.38 & 98.48 & 99.75 & 99.78 & 99.23 & 96.62 & - & 97.98 \\
\midrule
\multirow{10}{*}{MoCov3\cite{mocov3}} & 0 & - & 87.10 & 87.79 & 90.33 & 89.95 & 94.07 & 91.11 & 72.18 & 69.31 & 69.45 & 83.48 \\
& 1 & 93.05 & - & 98.12 & 97.41 & 98.20 & 98.26 & 99.03 & 94.47 & 93.38 & 75.58 & 94.17 \\
& 2 & 82.96 & 95.82 & - & 83.13 & 78.83 & 93.45 & 82.30 & 73.48 & 94.46 & 96.90 & 86.81 \\
& 3 & 91.08 & 93.10 & 90.56 & - & 74.01 & 74.04 & 84.32 & 79.70 & 93.94 & 93.13 & 85.99 \\
& 4 & 93.82 & 96.02 & 86.55 & 88.85 & - & 92.48 & 92.56 & 66.43 & 96.26 & 96.97 & 89.99 \\
& 5 & 94.32 & 95.45 & 91.58 & 74.17 & 88.34 & - & 94.31 & 69.63 & 97.36 & 98.02 & 89.24 \\
& 6 & 95.09 & 96.76 & 88.10 & 87.53 & 80.44 & 90.97 & - & 94.57 & 92.31 & 97.94 & 91.52 \\
& 7 & 91.62 & 96.60 & 92.38 & 91.56 & 80.46 & 92.20 & 96.91 & - & 92.62 & 95.95 & 92.25 \\
& 8 & 80.42 & 88.78 & 96.30 & 93.92 & 96.62 & 95.98 & 95.00 & 95.97 & - & 79.94 & 91.44 \\
& 9 & 92.33 & 79.04 & 97.95 & 97.19 & 98.14 & 98.28 & 98.73 & 97.74 & 88.23 & - & 94.18 \\
\midrule
\multirow{10}{*}{DINOv2\cite{dinov2}} & 0 & - & 60.94 & 65.14 & 64.79 & 71.95 & 62.40 & 76.86 & 61.40 & 39.99 & 51.08 & 61.62 \\
& 1 & 73.33 & - & 59.97 & 48.33 & 56.38 & 44.85 & 57.27 & 41.11 & 60.09 & 45.81 & 54.13 \\
& 2 & 69.07 & 55.66 & - & 49.75 & 44.54 & 46.53 & 51.42 & 43.47 & 57.03 & 53.23 & 52.30 \\
& 3 & 79.75 & 63.89 & 61.86 & - & 56.82 & 47.72 & 59.42 & 47.58 & 70.69 & 61.79 & 61.06 \\
& 4 & 81.90 & 68.04 & 59.55 & 59.30 & - & 57.66 & 54.34 & 53.14 & 73.53 & 68.04 & 63.95 \\
& 5 & 81.63 & 65.90 & 63.37 & 52.89 & 58.20 & - & 61.38 & 48.90 & 73.32 & 64.01 & 63.29 \\
& 6 & 86.28 & 71.81 & 62.40 & 57.83 & 51.69 & 56.97 & - & 54.87 & 81.90 & 74.10 & 66.43 \\
& 7 & 84.48 & 68.52 & 64.60 & 58.30 & 57.49 & 54.55 & 58.26 & - & 77.39 & 66.77 & 65.60 \\
& 8 & 64.17 & 71.12 & 75.01 & 73.31 & 79.28 & 71.02 & 84.92 & 71.20 & - & 61.16 & 72.35 \\
& 9 & 75.23 & 59.90 & 68.79 & 58.47 & 68.79 & 54.25 & 72.32 & 50.25 & 62.55 & - & 63.39 \\
\midrule
\multirow{10}{*}{BEiTv2\cite{beitv2}} & 0 & - & 94.83 & 97.42 & 99.31 & 97.01 & 99.79 & 99.34 & 99.44 & 89.14 & 91.32 & 96.40 \\
& 1 & 99.01 & - & 99.91 & 99.83 & 99.72 & 99.90 & 99.98 & 99.74 & 92.98 & 94.11 & 98.35 \\
& 2 & 85.99 & 99.23 & - & 98.03 & 81.97 & 99.25 & 93.81 & 93.23 & 96.98 & 99.43 & 94.21 \\
& 3 & 98.39 & 98.20 & 97.60 & - & 90.40 & 85.67 & 74.37 & 95.00 & 99.49 & 98.30 & 93.05 \\
& 4 & 99.57 & 99.62 & 98.82 & 88.31 & - & 99.31 & 99.03 & 77.26 & 99.32 & 99.88 & 95.68 \\
& 5 & 99.57 & 99.31 & 99.40 & 80.54 & 97.51 & - & 99.14 & 53.39 & 99.72 & 98.76 & 91.93 \\
& 6 & 99.46 & 99.75 & 97.00 & 96.55 & 97.07 & 99.04 & - & 99.35 & 99.71 & 99.91 & 98.65 \\
& 7 & 99.12 & 98.87 & 99.76 & 98.94 & 92.57 & 99.02 & 99.92 & - & 98.73 & 95.91 & 98.09 \\
& 8 & 88.49 & 93.17 & 99.83 & 99.74 & 99.82 & 99.94 & 99.94 & 99.86 & - & 96.15 & 97.44 \\
& 9 & 99.19 & 96.36 & 99.96 & 99.75 & 99.75 & 99.97 & 99.97 & 99.80 & 97.88 & - & 99.18 \\
\midrule
\multirow{10}{*}{BEiT\cite{beit} } & 0 & - & 95.65 & 95.16 & 98.75 & 94.10 & 99.60 & 98.60 & 98.70 & 83.00 & 85.22 & 94.31 \\
& 1 & 97.46 & - & 99.85 & 99.63 & 99.47 & 99.91 & 99.96 & 99.55 & 95.42 & 90.32 & 97.95 \\
& 2 & 96.98 & 98.44 & - & 89.21 & 77.98 & 98.19 & 97.53 & 72.68 & 97.65 & 99.17 & 91.98 \\
& 3 & 97.62 & 95.80 & 96.95 & - & 80.24 & 86.56 & 72.85 & 86.98 & 99.38 & 94.34 & 90.08 \\
& 4 & 99.40 & 98.45 & 98.57 & 98.12 & - & 98.92 & 96.00 & 70.81 & 98.42 & 93.32 & 94.67 \\
& 5 & 99.15 & 99.03 & 98.66 & 82.32 & 82.56 & - & 98.78 & 47.63 & 99.81 & 99.62 & 89.73 \\
& 6 & 99.32 & 99.56 & 96.90 & 96.87 & 93.96 & 99.06 & - & 99.56 & 99.78 & 99.89 & 98.32 \\
& 7 & 98.76 & 98.51 & 99.55 & 99.00 & 93.66 & 99.01 & 99.84 & - & 98.99 & 96.76 & 98.23 \\
& 8 & 92.28 & 92.62 & 99.44 & 99.40 & 99.72 & 99.80 & 99.81 & 99.44 & - & 92.60 & 97.24 \\
& 9 & 99.07 & 93.71 & 99.93 & 99.77 & 99.80 & 99.97 & 99.98 & 99.62 & 97.69 & - & 98.84 \\
\bottomrule
\end{tabular}
\caption{AUROC (\%) of one-class OOD Detection on CIFAR-10 \cite{cifar} using KL-Matching \cite{maxlogit}. Pretrained models include classification task \cite{vit}, MoCov3 \cite{mocov3}, DINOv2 \cite{dinov2}, BEiTv2 \cite{beitv2} and BEiT \cite{beit}.}
\label{tab:one-class-detailed-kl}
\end{table*}

% -------------------------------------------------------------
\begin{table*}[t]
\small
\centering
\setlength{\tabcolsep}{2mm}
\begin{tabular}{c|c|cccccccccc|c}
\toprule
\multirow{2}{*}{Models} & \multirow{2}{*}{ID class} & \multicolumn{10}{c|}{OOD class} & \multirow{2}{*}{Average} \\
 &  & 0 & 1 & 2 & 3 & 4 & 5 & 6 & 7 & 8 & 9 & \\
\midrule
\multirow{10}{*}{ViT\cite{vit}} & 0 & - & 88.70 & 96.73 & 99.02 & 96.98 & 99.82 & 98.77 & 98.21 & 66.32 & 79.11 & 91.52 \\
& 1 & 98.03 & - & 99.97 & 99.93 & 99.92 & 99.99 & 99.98 & 99.87 & 95.48 & 69.60 & 95.86 \\
& 2 & 96.51 & 99.92 & - & 94.67 & 75.60 & 98.06 & 89.95 & 90.72 & 99.08 & 99.83 & 93.82 \\
& 3 & 98.54 & 99.70 & 94.27 & - & 75.83 & 66.18 & 89.15 & 81.41 & 99.56 & 98.99 & 89.29 \\
& 4 & 99.25 & 99.85 & 93.29 & 95.01 & - & 95.08 & 95.51 & 70.54 & 99.40 & 99.54 & 94.16 \\
& 5 & 99.74 & 99.97 & 98.38 & 87.74 & 89.16 & - & 98.40 & 85.77 & 99.92 & 99.85 & 95.44 \\
& 6 & 99.44 & 99.96 & 93.81 & 95.33 & 88.84 & 98.00 & - & 97.18 & 99.79 & 99.90 & 96.92 \\
& 7 & 98.85 & 99.64 & 97.53 & 95.31 & 76.90 & 91.92 & 99.25 & - & 98.41 & 98.70 & 95.17 \\
& 8 & 90.11 & 89.74 & 99.59 & 99.72 & 99.56 & 99.96 & 99.60 & 99.65 & - & 85.70 & 95.96 \\
& 9 & 98.11 & 85.44 & 99.96 & 99.93 & 99.80 & 99.99 & 99.98 & 99.75 & 96.08 & - & 97.67 \\
\midrule
\multirow{10}{*}{MoCov3\cite{mocov3}} & 0 & - & 96.11 & 97.08 & 99.13 & 98.40 & 99.75 & 98.56 & 99.00 & 73.05 & 87.16 & 94.25 \\
& 1 & 96.02 & - & 99.79 & 99.76 & 99.76 & 99.90 & 99.82 & 99.71 & 92.94 & 68.77 & 95.16 \\
& 2 & 93.25 & 99.90 & - & 94.04 & 80.49 & 97.94 & 88.20 & 91.06 & 98.09 & 99.60 & 93.62 \\
& 3 & 97.15 & 99.60 & 93.31 & - & 79.89 & 71.78 & 87.59 & 86.66 & 98.02 & 98.72 & 90.30 \\
& 4 & 98.28 & 99.91 & 92.84 & 93.95 & - & 96.47 & 94.59 & 73.27 & 98.35 & 99.53 & 94.13 \\
& 5 & 99.12 & 99.90 & 97.62 & 84.00 & 89.42 & - & 98.45 & 89.13 & 99.38 & 99.45 & 95.16 \\
& 6 & 99.38 & 99.97 & 95.00 & 95.30 & 93.13 & 98.72 & - & 98.83 & 99.66 & 99.90 & 97.77 \\
& 7 & 97.23 & 99.70 & 96.70 & 95.15 & 76.84 & 93.22 & 98.94 & - & 97.06 & 98.53 & 94.82 \\
& 8 & 87.65 & 95.39 & 99.26 & 99.45 & 99.47 & 99.76 & 99.27 & 99.57 & - & 89.66 & 96.61 \\
& 9 & 96.24 & 86.59 & 99.84 & 99.83 & 99.84 & 99.91 & 99.88 & 99.77 & 92.68 & - & 97.18 \\
\midrule
\multirow{10}{*}{DINOv2\cite{dinov2}} & 0 & - & 72.83 & 64.15 & 68.05 & 65.47 & 68.53 & 68.80 & 70.10 & 54.59 & 69.72 & 66.92 \\
& 1 & 66.18 & - & 66.00 & 56.84 & 62.76 & 58.68 & 60.05 & 57.11 & 54.97 & 52.30 & 59.43 \\
& 2 & 66.58 & 77.06 & - & 57.10 & 45.49 & 56.78 & 49.57 & 62.65 & 67.99 & 76.34 & 62.17 \\
& 3 & 74.83 & 77.23 & 59.35 & - & 52.76 & 50.60 & 51.10 & 63.03 & 71.63 & 76.36 & 64.10 \\
& 4 & 82.00 & 86.75 & 63.82 & 67.27 & - & 66.91 & 57.16 & 71.45 & 80.04 & 85.45 & 73.43 \\
& 5 & 79.01 & 80.38 & 61.21 & 54.33 & 54.49 & - & 54.53 & 64.44 & 75.72 & 80.20 & 67.15 \\
& 6 & 85.87 & 86.34 & 66.60 & 66.63 & 56.40 & 66.99 & - & 75.17 & 84.70 & 87.00 & 75.08 \\
& 7 & 76.76 & 76.51 & 59.87 & 56.22 & 49.64 & 53.88 & 52.70 & - & 73.05 & 72.37 & 63.44 \\
& 8 & 62.96 & 74.15 & 73.59 & 72.77 & 74.25 & 72.95 & 77.20 & 76.44 & - & 71.13 & 72.83 \\
& 9 & 68.96 & 59.80 & 70.76 & 61.83 & 68.02 & 62.95 & 66.87 & 58.87 & 57.35 & - & 63.94 \\
\midrule
\multirow{10}{*}{BEiTv2\cite{beitv2}} & 0 & - & 98.30 & 99.41 & 99.91 & 99.60 & 99.98 & 99.86 & 99.68 & 91.03 & 93.81 & 97.95 \\
& 1 & 99.66 & - & 100.00 & 99.99 & 99.99 & 99.99 & 100.00 & 99.99 & 98.49 & 78.72 & 97.43 \\
& 2 & 97.60 & 99.96 & - & 99.17 & 93.13 & 99.48 & 97.72 & 98.58 & 99.58 & 99.92 & 98.35 \\
& 3 & 99.74 & 99.87 & 98.56 & - & 94.77 & 80.80 & 94.73 & 97.37 & 99.89 & 99.90 & 96.18 \\
& 4 & 99.83 & 99.90 & 98.68 & 99.06 & - & 99.14 & 99.16 & 87.86 & 99.83 & 99.88 & 98.15 \\
& 5 & 99.94 & 99.97 & 99.62 & 91.48 & 98.24 & - & 99.70 & 96.71 & 99.95 & 99.81 & 98.38 \\
& 6 & 99.89 & 99.99 & 99.39 & 99.25 & 99.14 & 99.68 & - & 99.83 & 99.94 & 99.98 & 99.68 \\
& 7 & 99.67 & 99.82 & 99.76 & 99.75 & 97.38 & 99.45 & 99.98 & - & 99.67 & 99.66 & 99.46 \\
& 8 & 97.13 & 98.60 & 99.92 & 99.98 & 99.95 & 99.97 & 99.98 & 99.97 & - & 95.87 & 99.04 \\
& 9 & 99.36 & 95.53 & 100.00 & 99.99 & 100.00 & 100.00 & 100.00 & 99.99 & 98.54 & - & 99.27 \\
\midrule
\multirow{10}{*}{BEiT\cite{beit} } & 0 & - & 98.74 & 99.39 & 99.91 & 99.59 & 99.96 & 99.83 & 99.77 & 91.23 & 93.39 & 97.98 \\
& 1 & 99.32 & - & 100.00 & 99.98 & 99.98 & 100.00 & 100.00 & 99.97 & 99.02 & 88.68 & 98.55 \\
& 2 & 99.25 & 99.97 & - & 98.32 & 90.25 & 99.21 & 97.47 & 98.08 & 99.79 & 99.88 & 98.02 \\
& 3 & 99.78 & 99.85 & 99.09 & - & 94.53 & 83.32 & 94.51 & 97.40 & 99.94 & 99.60 & 96.45 \\
& 4 & 99.88 & 99.90 & 98.82 & 98.63 & - & 98.89 & 98.95 & 93.34 & 99.88 & 99.78 & 98.67 \\
& 5 & 99.98 & 99.88 & 99.77 & 90.61 & 97.43 & - & 99.60 & 93.86 & 99.99 & 99.83 & 97.88 \\
& 6 & 99.95 & 100.00 & 99.28 & 99.13 & 98.97 & 99.67 & - & 99.97 & 99.99 & 99.99 & 99.66 \\
& 7 & 99.55 & 99.45 & 99.76 & 99.41 & 97.82 & 99.04 & 99.94 & - & 99.49 & 99.13 & 99.29 \\
& 8 & 95.97 & 97.99 & 99.95 & 99.95 & 99.90 & 99.98 & 99.98 & 99.96 & - & 96.03 & 98.86 \\
& 9 & 99.38 & 94.73 & 100.00 & 99.98 & 99.98 & 100.00 & 100.00 & 99.85 & 98.85 & - & 99.20 \\
\bottomrule
\end{tabular}
\caption{AUROC (\%) of one-class OOD Detection on CIFAR-10 \cite{cifar} using Residual\cite{vim}. Pretrained models include classification task \cite{vit}, MoCov3 \cite{mocov3}, DINOv2 \cite{dinov2}, BEiTv2 \cite{beitv2} and BEiT \cite{beit}.}
\label{tab:one-class-detailed-residual}
\end{table*}

% -------------------------------------------------------------
\begin{table*}[t]
\small
\centering
\setlength{\tabcolsep}{2mm}
\begin{tabular}{c|c|cccccccccc|c}
\toprule
\multirow{2}{*}{Models} & \multirow{2}{*}{ID class} & \multicolumn{10}{c|}{OOD class} & \multirow{2}{*}{Average} \\
 &  & 0 & 1 & 2 & 3 & 4 & 5 & 6 & 7 & 8 & 9 & \\
\midrule
\multirow{10}{*}{ViT\cite{vit}} & 0 & - & 89.17 & 95.53 & 98.62 & 95.81 & 99.70 & 97.89 & 97.15 & 65.74 & 78.81 & 90.94 \\
& 1 & 98.20 & - & 99.94 & 99.91 & 99.86 & 99.98 & 99.96 & 99.79 & 96.03 & 71.41 & 96.12 \\
& 2 & 96.19 & 99.84 & - & 94.76 & 77.37 & 98.03 & 89.80 & 90.19 & 98.81 & 99.68 & 93.85 \\
& 3 & 98.08 & 99.40 & 93.07 & - & 75.92 & 68.55 & 87.75 & 81.75 & 99.21 & 98.35 & 89.12 \\
& 4 & 99.19 & 99.81 & 93.59 & 95.16 & - & 96.03 & 95.52 & 70.51 & 99.35 & 99.54 & 94.30 \\
& 5 & 99.65 & 99.91 & 98.30 & 88.92 & 91.23 & - & 98.31 & 87.68 & 99.86 & 99.72 & 95.95 \\
& 6 & 99.30 & 99.93 & 93.91 & 95.23 & 89.50 & 98.04 & - & 96.95 & 99.64 & 99.82 & 96.93 \\
& 7 & 98.62 & 99.50 & 97.22 & 95.53 & 77.46 & 92.86 & 99.03 & - & 98.20 & 98.59 & 95.22 \\
& 8 & 90.26 & 90.45 & 99.48 & 99.67 & 99.47 & 99.93 & 99.48 & 99.57 & - & 87.26 & 96.18 \\
& 9 & 98.16 & 86.56 & 99.93 & 99.91 & 99.76 & 99.99 & 99.97 & 99.70 & 96.39 & - & 97.82 \\
\midrule
\multirow{10}{*}{MoCov3\cite{mocov3}} & 0 & - & 95.82 & 95.69 & 98.75 & 97.49 & 99.58 & 97.83 & 98.04 & 71.17 & 85.75 & 93.35 \\
& 1 & 95.59 & - & 99.66 & 99.69 & 99.67 & 99.85 & 99.74 & 99.50 & 92.72 & 68.79 & 95.02 \\
& 2 & 92.17 & 99.77 & - & 93.66 & 80.97 & 97.58 & 87.61 & 90.67 & 97.16 & 99.21 & 93.20 \\
& 3 & 95.87 & 99.33 & 91.17 & - & 78.08 & 71.77 & 86.00 & 84.32 & 96.86 & 98.04 & 89.05 \\
& 4 & 97.97 & 99.79 & 92.35 & 94.00 & - & 96.40 & 94.38 & 73.76 & 98.11 & 99.26 & 94.00 \\
& 5 & 98.66 & 99.78 & 97.00 & 84.31 & 90.24 & - & 97.96 & 89.13 & 98.93 & 99.14 & 95.02 \\
& 6 & 99.13 & 99.94 & 94.33 & 95.19 & 92.56 & 98.56 & - & 98.38 & 99.45 & 99.82 & 97.48 \\
& 7 & 96.61 & 99.55 & 96.01 & 94.83 & 76.00 & 92.97 & 98.57 & - & 96.47 & 98.15 & 94.35 \\
& 8 & 87.08 & 95.47 & 98.86 & 99.27 & 99.20 & 99.62 & 98.97 & 99.20 & - & 89.75 & 96.38 \\
& 9 & 95.98 & 86.94 & 99.74 & 99.77 & 99.76 & 99.87 & 99.82 & 99.63 & 92.63 & - & 97.13 \\
\midrule
\multirow{10}{*}{DINOv2\cite{dinov2}} & 0 & - & 73.26 & 64.39 & 68.07 & 65.83 & 68.69 & 69.50 & 70.14 & 54.72 & 70.07 & 67.19 \\
& 1 & 66.64 & - & 66.59 & 57.46 & 63.19 & 58.81 & 60.71 & 57.10 & 55.24 & 52.26 & 59.78 \\
& 2 & 67.18 & 77.06 & - & 57.44 & 45.86 & 57.11 & 49.82 & 63.01 & 68.56 & 76.62 & 62.52 \\
& 3 & 75.38 & 77.23 & 59.51 & - & 52.41 & 50.47 & 50.53 & 63.05 & 72.42 & 76.62 & 64.18 \\
& 4 & 82.15 & 86.71 & 63.94 & 67.52 & - & 67.12 & 57.42 & 71.56 & 80.55 & 86.07 & 73.67 \\
& 5 & 77.37 & 79.22 & 60.55 & 53.76 & 54.45 & - & 54.10 & 63.90 & 75.10 & 78.91 & 66.37 \\
& 6 & 86.06 & 86.59 & 66.51 & 66.79 & 56.39 & 67.45 & - & 75.36 & 84.95 & 87.54 & 75.29 \\
& 7 & 77.61 & 76.62 & 59.83 & 56.58 & 49.58 & 54.03 & 53.04 & - & 74.18 & 72.88 & 63.82 \\
& 8 & 63.06 & 74.53 & 74.59 & 73.16 & 75.12 & 73.67 & 78.29 & 76.87 & - & 71.34 & 73.40 \\
& 9 & 69.38 & 60.16 & 70.57 & 62.54 & 67.91 & 63.32 & 67.09 & 58.82 & 58.01 & - & 64.20 \\
\midrule
\multirow{10}{*}{BEiTv2\cite{beitv2}} & 0 & - & 98.36 & 99.38 & 99.94 & 99.70 & 99.99 & 99.90 & 99.70 & 91.92 & 94.07 & 98.11 \\
& 1 & 99.69 & - & 100.00 & 99.99 & 99.99 & 100.00 & 100.00 & 99.99 & 98.96 & 80.71 & 97.70 \\
& 2 & 97.78 & 99.94 & - & 99.37 & 94.03 & 99.54 & 98.04 & 98.32 & 99.54 & 99.88 & 98.49 \\
& 3 & 99.72 & 99.89 & 98.56 & - & 94.80 & 81.26 & 95.70 & 96.85 & 99.85 & 99.87 & 96.28 \\
& 4 & 99.79 & 99.92 & 98.69 & 99.20 & - & 99.23 & 99.34 & 88.58 & 99.84 & 99.88 & 98.27 \\
& 5 & 99.94 & 99.98 & 99.57 & 92.56 & 98.39 & - & 99.73 & 97.18 & 99.92 & 99.85 & 98.57 \\
& 6 & 99.87 & 100.00 & 99.44 & 99.37 & 99.25 & 99.71 & - & 99.80 & 99.92 & 99.96 & 99.70 \\
& 7 & 99.68 & 99.80 & 99.74 & 99.77 & 97.64 & 99.50 & 99.96 & - & 99.69 & 99.62 & 99.49 \\
& 8 & 97.51 & 98.74 & 99.90 & 99.97 & 99.92 & 99.95 & 99.95 & 99.91 & - & 96.76 & 99.18 \\
& 9 & 99.41 & 95.56 & 100.00 & 99.99 & 99.99 & 100.00 & 100.00 & 99.98 & 98.60 & - & 99.28 \\
\midrule
\multirow{10}{*}{BEiT\cite{beit} } & 0 & - & 98.65 & 99.41 & 99.91 & 99.59 & 99.96 & 99.86 & 99.72 & 91.50 & 94.10 & 98.08 \\
& 1 & 99.45 & - & 100.00 & 99.98 & 99.98 & 100.00 & 100.00 & 99.96 & 99.14 & 89.24 & 98.64 \\
& 2 & 99.25 & 99.95 & - & 98.42 & 90.61 & 99.25 & 97.69 & 98.32 & 99.78 & 99.87 & 98.13 \\
& 3 & 99.77 & 99.80 & 99.09 & - & 95.07 & 84.11 & 94.70 & 97.80 & 99.94 & 99.63 & 96.66 \\
& 4 & 99.89 & 99.89 & 98.88 & 98.85 & - & 99.06 & 99.09 & 93.79 & 99.90 & 99.77 & 98.79 \\
& 5 & 99.98 & 99.83 & 99.75 & 91.54 & 97.74 & - & 99.61 & 95.89 & 99.99 & 99.79 & 98.23 \\
& 6 & 99.95 & 100.00 & 99.33 & 99.19 & 99.06 & 99.67 & - & 99.96 & 99.98 & 99.98 & 99.68 \\
& 7 & 99.61 & 99.48 & 99.78 & 99.46 & 97.85 & 99.14 & 99.94 & - & 99.53 & 99.18 & 99.33 \\
& 8 & 96.51 & 98.01 & 99.95 & 99.96 & 99.92 & 99.98 & 99.98 & 99.95 & - & 96.05 & 98.92 \\
& 9 & 99.47 & 94.84 & 100.00 & 99.99 & 99.98 & 100.00 & 100.00 & 99.90 & 98.97 & - & 99.24 \\
\bottomrule
\end{tabular}
\caption{AUROC (\%) of one-class OOD Detection on CIFAR-10 \cite{cifar} using Mahalanobis\cite{mahalanobis}. Pretrained models include classification task \cite{vit}, MoCov3 \cite{mocov3}, DINOv2 \cite{dinov2}, BEiTv2 \cite{beitv2} and BEiT \cite{beit}.}
\label{tab:one-class-detailed-distance}
\end{table*}

% -------------------------------------------------------------
\begin{table*}[t]
\small
\centering
\setlength{\tabcolsep}{2mm}
\begin{tabular}{c|c|cccccccccc|c}
\toprule
\multirow{2}{*}{Models} & \multirow{2}{*}{ID class} & \multicolumn{10}{c|}{OOD class} & \multirow{2}{*}{Average} \\
 &  & 0 & 1 & 2 & 3 & 4 & 5 & 6 & 7 & 8 & 9 & \\
\midrule
\multirow{10}{*}{ViT\cite{vit}} & 0 & - & 90.64 & 97.03 & 99.20 & 97.19 & 99.86 & 98.92 & 98.45 & 67.16 & 80.51 & 92.11 \\
& 1 & 97.97 & - & 99.98 & 99.93 & 99.93 & 99.99 & 99.99 & 99.89 & 95.22 & 67.49 & 95.60 \\
& 2 & 96.31 & 99.93 & - & 94.88 & 74.69 & 98.10 & 89.81 & 90.86 & 99.10 & 99.85 & 93.72 \\
& 3 & 98.61 & 99.76 & 94.39 & - & 74.04 & 64.74 & 88.69 & 81.14 & 99.59 & 99.07 & 88.89 \\
& 4 & 99.25 & 99.86 & 93.62 & 95.27 & - & 95.28 & 95.60 & 71.21 & 99.41 & 99.54 & 94.34 \\
& 5 & 99.75 & 99.98 & 98.47 & 88.47 & 89.05 & - & 98.42 & 86.11 & 99.93 & 99.87 & 95.56 \\
& 6 & 99.46 & 99.97 & 94.05 & 95.60 & 88.80 & 98.08 & - & 97.32 & 99.80 & 99.91 & 97.00 \\
& 7 & 98.90 & 99.69 & 97.61 & 95.49 & 75.97 & 91.93 & 99.27 & - & 98.47 & 98.79 & 95.12 \\
& 8 & 90.10 & 90.70 & 99.63 & 99.77 & 99.59 & 99.97 & 99.63 & 99.68 & - & 86.14 & 96.14 \\
& 9 & 98.12 & 86.67 & 99.96 & 99.93 & 99.79 & 99.99 & 99.99 & 99.75 & 96.03 & - & 97.80 \\
\midrule
\multirow{10}{*}{MoCov3\cite{mocov3}} & 0 & - & 97.10 & 97.66 & 99.36 & 98.74 & 99.83 & 98.88 & 99.27 & 74.06 & 88.58 & 94.83 \\
& 1 & 96.25 & - & 99.82 & 99.79 & 99.79 & 99.91 & 99.84 & 99.74 & 93.06 & 67.99 & 95.13 \\
& 2 & 93.80 & 99.93 & - & 94.63 & 80.47 & 98.24 & 88.49 & 91.63 & 98.37 & 99.69 & 93.92 \\
& 3 & 97.60 & 99.68 & 94.10 & - & 79.97 & 72.24 & 87.98 & 87.63 & 98.32 & 98.96 & 90.72 \\
& 4 & 98.49 & 99.93 & 93.56 & 94.53 & - & 96.84 & 94.90 & 74.27 & 98.52 & 99.61 & 94.52 \\
& 5 & 99.29 & 99.93 & 97.94 & 85.08 & 89.84 & - & 98.61 & 89.87 & 99.49 & 99.57 & 95.51 \\
& 6 & 99.46 & 99.97 & 95.39 & 95.71 & 93.39 & 98.87 & - & 98.97 & 99.69 & 99.90 & 97.93 \\
& 7 & 97.57 & 99.77 & 97.08 & 95.65 & 76.65 & 93.74 & 99.06 & - & 97.39 & 98.74 & 95.07 \\
& 8 & 88.58 & 96.14 & 99.39 & 99.55 & 99.55 & 99.81 & 99.39 & 99.64 & - & 90.57 & 96.96 \\
& 9 & 96.55 & 87.81 & 99.87 & 99.86 & 99.86 & 99.93 & 99.90 & 99.79 & 92.96 & - & 97.39 \\
\midrule
\multirow{10}{*}{DINOv2\cite{dinov2}} & 0 & - & 72.84 & 64.24 & 68.20 & 65.56 & 68.69 & 68.90 & 70.20 & 54.60 & 69.76 & 67.00 \\
& 1 & 66.30 & - & 66.15 & 57.07 & 62.92 & 58.93 & 60.22 & 57.30 & 55.07 & 52.37 & 59.59 \\
& 2 & 66.61 & 77.05 & - & 57.20 & 45.49 & 56.89 & 49.57 & 62.70 & 68.01 & 76.37 & 62.21 \\
& 3 & 74.84 & 77.16 & 59.30 & - & 52.69 & 50.62 & 51.02 & 63.03 & 71.61 & 76.33 & 64.06 \\
& 4 & 82.04 & 86.74 & 63.85 & 67.37 & - & 67.00 & 57.18 & 71.49 & 80.07 & 85.48 & 73.47 \\
& 5 & 79.00 & 80.30 & 61.17 & 54.32 & 54.41 & - & 54.44 & 64.39 & 75.69 & 80.16 & 67.10 \\
& 6 & 85.90 & 86.33 & 66.63 & 66.71 & 56.41 & 67.08 & - & 75.22 & 84.73 & 87.03 & 75.12 \\
& 7 & 76.80 & 76.47 & 59.90 & 56.30 & 49.63 & 53.98 & 52.69 & - & 73.07 & 72.36 & 63.47 \\
& 8 & 63.00 & 74.18 & 73.68 & 72.92 & 74.35 & 73.11 & 77.30 & 76.54 & - & 71.19 & 72.92 \\
& 9 & 69.06 & 59.79 & 70.88 & 62.04 & 68.16 & 63.16 & 67.02 & 59.00 & 57.41 & - & 64.06 \\
\midrule
\multirow{10}{*}{BEiTv2\cite{beitv2}} & 0 & - & 97.86 & 99.24 & 99.90 & 99.48 & 99.97 & 99.73 & 99.61 & 87.59 & 89.06 & 96.94 \\
& 1 & 99.49 & - & 99.99 & 99.99 & 99.98 & 99.98 & 100.00 & 99.99 & 97.62 & 66.23 & 95.92 \\
& 2 & 95.66 & 99.95 & - & 98.84 & 90.50 & 98.98 & 95.46 & 97.71 & 99.07 & 99.77 & 97.33 \\
& 3 & 99.44 & 99.79 & 98.00 & - & 93.19 & 72.65 & 90.51 & 96.39 & 99.65 & 99.32 & 94.33 \\
& 4 & 99.72 & 99.87 & 98.37 & 98.87 & - & 98.76 & 98.78 & 84.32 & 99.72 & 99.75 & 97.57 \\
& 5 & 99.86 & 99.95 & 99.47 & 91.21 & 97.83 & - & 99.52 & 95.97 & 99.88 & 99.49 & 98.13 \\
& 6 & 99.85 & 99.99 & 99.26 & 99.14 & 98.93 & 99.60 & - & 99.81 & 99.91 & 99.92 & 99.60 \\
& 7 & 99.55 & 99.78 & 99.68 & 99.72 & 96.47 & 99.17 & 99.96 & - & 99.45 & 99.34 & 99.24 \\
& 8 & 96.02 & 98.31 & 99.90 & 99.97 & 99.94 & 99.96 & 99.96 & 99.97 & - & 93.04 & 98.56 \\
& 9 & 99.17 & 95.57 & 100.00 & 99.99 & 99.99 & 99.99 & 100.00 & 99.99 & 98.21 & - & 99.21 \\
\midrule
\multirow{10}{*}{BEiT\cite{beit} } & 0 & - & 98.76 & 99.35 & 99.90 & 99.48 & 99.95 & 99.76 & 99.78 & 88.74 & 91.93 & 97.52 \\
& 1 & 99.15 & - & 100.00 & 99.98 & 99.97 & 100.00 & 100.00 & 99.97 & 98.78 & 84.81 & 98.07 \\
& 2 & 99.12 & 99.98 & - & 97.89 & 88.04 & 98.88 & 96.49 & 97.93 & 99.73 & 99.85 & 97.55 \\
& 3 & 99.77 & 99.83 & 98.95 & - & 93.35 & 78.82 & 92.42 & 97.56 & 99.94 & 99.38 & 95.56 \\
& 4 & 99.87 & 99.90 & 98.65 & 98.48 & - & 98.66 & 98.75 & 93.22 & 99.85 & 99.73 & 98.57 \\
& 5 & 99.97 & 99.88 & 99.73 & 90.68 & 97.10 & - & 99.49 & 94.34 & 99.99 & 99.76 & 97.88 \\
& 6 & 99.96 & 100.00 & 99.27 & 99.08 & 98.83 & 99.64 & - & 99.98 & 99.99 & 99.99 & 99.64 \\
& 7 & 99.53 & 99.46 & 99.75 & 99.39 & 97.50 & 98.89 & 99.92 & - & 99.46 & 99.03 & 99.22 \\
& 8 & 95.53 & 98.13 & 99.94 & 99.95 & 99.87 & 99.97 & 99.98 & 99.96 & - & 95.62 & 98.77 \\
& 9 & 99.32 & 95.30 & 100.00 & 99.98 & 99.97 & 100.00 & 100.00 & 99.86 & 98.67 & - & 99.23 \\
\bottomrule
\end{tabular}
\caption{AUROC (\%) of one-class OOD Detection on CIFAR-10 \cite{cifar} using ViM\cite{vim}. Pretrained models include classification task \cite{vit}, MoCov3 \cite{mocov3}, DINOv2 \cite{dinov2}, BEiTv2 \cite{beitv2} and BEiT \cite{beit}.}
\label{tab:one-class-detailed-vim}
\end{table*}

\end{document}